\documentclass{article}

\usepackage{iftex}
\ifPDFTeX
  \usepackage[utf8]{inputenc}
\fi
\usepackage[T1]{fontenc}
\PassOptionsToPackage{round}{natbib}
\usepackage{genai4health_2026}
\usepackage{hyperref}
\hypersetup{hidelinks}
\usepackage{url}
\usepackage{booktabs}
\usepackage{graphicx}
\usepackage{amsmath}
\usepackage{amssymb}
\usepackage{xcolor}
\usepackage{microtype}
\usepackage{tikz}

\usetikzlibrary{arrows.meta}

\graphicspath{{figures/}{auto/}}

\newcommand{\AUCAnatomyTwoEpFifty}{0.8654}

\newcommand{\AUCAncestor}{0.8487}
\newcommand{\AUCCoverEpFifty}{0.8643}

\newcommand{\AUCEnvelopeEpFifty}{0.8761}
\newcommand{\AUCEnvelopeEpHundred}{0.8807}
\newcommand{\AUCEnvelopeEpSeventyFive}{0.8803}
\newcommand{\AUCOracleEpFifty}{0.8740}
\newcommand{\AUCOracleEpHundred}{0.8855}
\newcommand{\AUCOracleEpSeventyFive}{0.8836}
\newcommand{\AUCRandomEpFifty}{0.8641}

\newcommand{\AUCRandomEpHundred}{0.8746}
\newcommand{\AUCRandomEpSeventyFive}{0.8723}

\newcommand{\CoverPeakEpoch}{73}
\newcommand{\DAnatomyOneEnvelopeEpThirty}{+0.44}
\newcommand{\DAnatomyOneEnvelopeEpThirtyCI}{[+0.09,\,+0.78]}

\newcommand{\DAnatomyTwoEnvelopeEpFifty}{-1.07}
\newcommand{\DAnatomyTwoEnvelopeEpFiftyCI}{[-1.67,\,-0.46]}

\newcommand{\DAnatomyTwoRandomEpFifty}{+0.13}

\newcommand{\DCoverRandomEpFifty}{+0.02}
\newcommand{\DCoverRandomEpFiftyCI}{[-0.50,\,+0.53]}

\newcommand{\DCoverRandomEpHundred}{-1.68}
\newcommand{\DCoverRandomEpHundredCI}{[-2.33,\,-1.05]}

\newcommand{\DEnvelopeRandomEpFifty}{+1.20}
\newcommand{\DEnvelopeRandomEpFiftyCI}{[+0.68,\,+1.73]}
\newcommand{\DEnvelopeRandomEpFiftyP}{$<$0.0001}
\newcommand{\DEnvelopeRandomEpFiftyQ}{$<$0.0001}
\newcommand{\DEnvelopeRandomEpHundred}{+0.62}
\newcommand{\DEnvelopeRandomEpHundredCI}{[+0.10,\,+1.13]}
\newcommand{\DEnvelopeRandomEpHundredP}{0.0199}
\newcommand{\DEnvelopeRandomEpHundredQ}{0.0299}
\newcommand{\DEnvelopeRandomEpSeventyFive}{+0.80}
\newcommand{\DEnvelopeRandomEpSeventyFiveCI}{[+0.28,\,+1.32]}
\newcommand{\DEnvelopeRandomEpSeventyFiveP}{0.0025}
\newcommand{\DEnvelopeRandomEpSeventyFiveQ}{0.0045}
\newcommand{\DOracleEnvelopeEpFifty}{-0.20}
\newcommand{\DOracleEnvelopeEpFiftyCI}{[-0.69,\,+0.29]}
\newcommand{\DOracleEnvelopeEpFiftyP}{0.4114}
\newcommand{\DOracleEnvelopeEpFiftyQ}{0.4114}
\newcommand{\DOracleEnvelopeEpHundred}{+0.47}
\newcommand{\DOracleEnvelopeEpHundredCI}{[+0.04,\,+0.91]}
\newcommand{\DOracleEnvelopeEpHundredP}{0.0294}
\newcommand{\DOracleEnvelopeEpHundredQ}{0.0378}
\newcommand{\DOracleEnvelopeEpSeventyFive}{+0.33}
\newcommand{\DOracleEnvelopeEpSeventyFiveCI}{[-0.13,\,+0.79]}
\newcommand{\DOracleEnvelopeEpSeventyFiveP}{0.1491}
\newcommand{\DOracleEnvelopeEpSeventyFiveQ}{0.1677}
\newcommand{\DOracleRandomEpFifty}{+0.99}
\newcommand{\DOracleRandomEpFiftyCI}{[+0.51,\,+1.49]}
\newcommand{\DOracleRandomEpFiftyP}{$<$0.0001}
\newcommand{\DOracleRandomEpFiftyQ}{0.0001}
\newcommand{\DOracleRandomEpHundred}{+1.09}
\newcommand{\DOracleRandomEpHundredCI}{[+0.57,\,+1.60]}
\newcommand{\DOracleRandomEpHundredP}{$<$0.0001}
\newcommand{\DOracleRandomEpHundredQ}{$<$0.0001}
\newcommand{\DOracleRandomEpSeventyFive}{+1.13}
\newcommand{\DOracleRandomEpSeventyFiveCI}{[+0.63,\,+1.64]}
\newcommand{\DOracleRandomEpSeventyFiveP}{$<$0.0001}
\newcommand{\DOracleRandomEpSeventyFiveQ}{$<$0.0001}

\newcommand{\Nboot}{10{,}000}

\newcommand{\Npos}{1466}

\newcommand{\Ntest}{3000}

\newcommand{\TAUCAnatomyTwoEpFifty}{0.8654}
\newcommand{\TAUCAnatomyTwoEpHundred}{---}
\newcommand{\TAUCAnatomyTwoEpSeventyFive}{0.8612}
\newcommand{\TAUCCoverEpFifty}{0.8643}
\newcommand{\TAUCCoverEpHundred}{0.8577}
\newcommand{\TAUCCoverEpSeventyFive}{0.8639}

\newcommand{\TAnatomyTwoRandomEpHundred}{---}
\newcommand{\TAnatomyTwoRandomEpSeventyFive}{$-1.11$}

\newcommand{\TCoverRandomEpHundred}{$-1.68$}
\newcommand{\TCoverRandomEpSeventyFive}{$-0.84$}

\newcommand{\DTFullImages}{576}
\newcommand{\DTValidGuides}{573}

\newcommand{\DTLegacyFloorMiss}{358}
\newcommand{\DTLegacyTruncation}{300}
\newcommand{\DTLegacyZeroContext}{55}
\newcommand{\DTLegacyScoredMass}{78.51}
\newcommand{\DTLegacyDeliveredMass}{74.57}
\newcommand{\DTLegacyContextMean}{10.13}

\newcommand{\CRHidRandom}{53.2}
\newcommand{\CRPurRandom}{30.9}
\newcommand{\CRUniqRandom}{113.1}
\newcommand{\CRSlotsRandom}{160.2}
\newcommand{\CRCtxRandom}{27.5}
\newcommand{\CRCtxSDRandom}{0.1}

\newcommand{\CRImgHidRandom}{53.0}

\newcommand{\CRImgPurRandom}{30.9}

\newcommand{\CRImgMaskRandom}{44.0}

\newcommand{\CRImgCtxRandom}{42.1}

\newcommand{\CRImgSlotsRandom}{159.3}

\newcommand{\CRHidCentroid}{59.7}
\newcommand{\CRPurCentroid}{38.2}
\newcommand{\CRUniqCentroid}{102.5}
\newcommand{\CRSlotsCentroid}{160.2}
\newcommand{\CRCtxCentroid}{31.5}
\newcommand{\CRCtxSDCentroid}{0.4}

\newcommand{\CRImgHidCentroid}{59.7}

\newcommand{\CRImgPurCentroid}{38.3}

\newcommand{\CRImgMaskCentroid}{39.9}

\newcommand{\CRImgCtxCentroid}{45.7}

\newcommand{\CRImgSlotsCentroid}{159.3}

\newcommand{\CRHidEnvelope}{70.7}
\newcommand{\CRPurEnvelope}{42.1}
\newcommand{\CRUniqEnvelope}{110.2}
\newcommand{\CRSlotsEnvelope}{160.2}
\newcommand{\CRCtxEnvelope}{29.2}
\newcommand{\CRCtxSDEnvelope}{0.5}

\newcommand{\CRImgHidEnvelope}{70.5}

\newcommand{\CRImgPurEnvelope}{42.1}

\newcommand{\CRImgMaskEnvelope}{42.9}

\newcommand{\CRImgCtxEnvelope}{43.7}

\newcommand{\CRImgSlotsEnvelope}{159.3}

\newcommand{\CRHidCover}{73.3}
\newcommand{\CRPurCover}{40.1}
\newcommand{\CRUniqCover}{120.1}
\newcommand{\CRSlotsCover}{160.2}
\newcommand{\CRCtxCover}{25.2}
\newcommand{\CRCtxSDCover}{0.4}

\newcommand{\CRImgHidCover}{73.2}

\newcommand{\CRImgPurCover}{40.1}

\newcommand{\CRImgMaskCover}{46.7}

\newcommand{\CRImgCtxCover}{39.9}

\newcommand{\CRImgSlotsCover}{159.3}

\newcommand{\CRHidAnatomyTwo}{79.7}
\newcommand{\CRPurAnatomyTwo}{96.9}
\newcommand{\CRUniqAnatomyTwo}{54.0}
\newcommand{\CRSlotsAnatomyTwo}{64.0}
\newcommand{\CRCtxAnatomyTwo}{63.3}
\newcommand{\CRCtxSDAnatomyTwo}{0.4}

\newcommand{\CRImgHidAnatomyTwo}{79.7}

\newcommand{\CRImgPurAnatomyTwo}{96.8}

\newcommand{\CRImgMaskAnatomyTwo}{21.1}

\newcommand{\CRImgCtxAnatomyTwo}{68.0}

\newcommand{\CRImgSlotsAnatomyTwo}{64.0}

\newcommand{\CRCtxGapCentroid}{4.0}
\newcommand{\CRCtxGapCentroidCI}{3.6--4.4}

\newcommand{\CRCtxGapEnvelope}{1.6}
\newcommand{\CRCtxGapEnvelopeCI}{1.1--2.1}

\newcommand{\CRAuditSeedCount}{3}
\newcommand{\CRAuditDraws}{100}
\newcommand{\CRAuditBatch}{64}
\newcommand{\CRAuditViews}{6{,}400}
\newcommand{\CRAuditVolumes}{256}
\newcommand{\CRAuditSlices}{25}
\newcommand{\CRImgDraws}{6{,}400}
\newcommand{\CRSamplerCommit}{86ae8823f3afc9fa2c9ed6a3cb0c459b513ba1c5}

\newcommand{\CRBandRows}{7}
\newcommand{\CRBandCols}{10}
\newcommand{\CRBandRegion}{0.28}
\newcommand{\CRBandLateral}{0.6}
\newcommand{\CRGuideVolumes}{300}
\newcommand{\CRGuideBscans}{3{,}000}
\newcommand{\CRGuidePerStratum}{50}
\newcommand{\CRCensusVolumes}{6{,}000}
\newcommand{\CRFOne}{4.8}
\newcommand{\CRFOneCI}{3.7--5.9}
\newcommand{\CRFOneCensus}{5.3}
\newcommand{\CRFOneNormal}{2.6}
\newcommand{\CRFOneGlaucoma}{6.5}
\newcommand{\CRFOneLowSNR}{7.5}
\newcommand{\CRFOneHighSNR}{2.3}
\newcommand{\CRFTwo}{0.9}
\newcommand{\CRIoUMedian}{0.41}
\newcommand{\CREnvAnyFlag}{3.8}
\newcommand{\CREnvAnyFlagCI}{2.7--5.0}
\newcommand{\CREnvFragmented}{1.5}
\newcommand{\CREnvThickness}{2.5}
\newcommand{\CREnvPosition}{0.4}
\newcommand{\CRMirageDice}{0.925}
\newcommand{\CRFairVisionAUC}{0.8649}

\newcommand{\RepEndpoint}{50}
\newcommand{\RepNBoot}{10{,}000}
\newcommand{\RepAUCCentroidSeedA}{0.8721}
\newcommand{\RepValAUCCentroidSeedA}{0.8547}
\newcommand{\RepHeadSDCentroidSeedA}{0.0007}
\newcommand{\RepAUCCentroidSeedB}{0.8754}
\newcommand{\RepValAUCCentroidSeedB}{0.8566}
\newcommand{\RepHeadSDCentroidSeedB}{0.0005}
\newcommand{\RepAUCRandomCBSeedA}{0.8650}
\newcommand{\RepValAUCRandomCBSeedA}{0.8463}
\newcommand{\RepHeadSDRandomCBSeedA}{0.0004}
\newcommand{\RepAUCEnvelopeSeedA}{0.8716}
\newcommand{\RepValAUCEnvelopeSeedA}{0.8525}
\newcommand{\RepHeadSDEnvelopeSeedA}{0.0002}
\newcommand{\RepAUCEnvelopeSeedB}{0.8710}
\newcommand{\RepValAUCEnvelopeSeedB}{0.8593}
\newcommand{\RepHeadSDEnvelopeSeedB}{0.0011}
\newcommand{\RepAUCRandomSeedA}{0.8624}
\newcommand{\RepValAUCRandomSeedA}{0.8495}
\newcommand{\RepHeadSDRandomSeedA}{0.0005}
\newcommand{\RepAUCRandomSeedB}{0.8635}
\newcommand{\RepValAUCRandomSeedB}{0.8443}
\newcommand{\RepHeadSDRandomSeedB}{0.0008}
\newcommand{\RepAUCCentroidOrig}{0.8752}
\newcommand{\RepValAUCCentroidOrig}{0.8560}
\newcommand{\RepHeadSDCentroidOrig}{0.0006}
\newcommand{\RepAUCEnvelopeOrig}{0.8755}
\newcommand{\RepValAUCEnvelopeOrig}{0.8597}
\newcommand{\RepHeadSDEnvelopeOrig}{0.0005}
\newcommand{\RepAUCRandomOrig}{0.8641}
\newcommand{\RepValAUCRandomOrig}{0.8467}
\newcommand{\RepHeadSDRandomOrig}{0.0005}
\newcommand{\RepHeadSDMax}{0.0011}
\newcommand{\RepDeltaCentroidSeedA}{+0.96}
\newcommand{\RepDeltaCentroidSeedACI}{[+0.52,\,+1.42]}
\newcommand{\RepValDeltaCentroidSeedA}{+0.52}
\newcommand{\RepDeltaCentroidSeedB}{+1.20}
\newcommand{\RepDeltaCentroidSeedBCI}{[+0.77,\,+1.64]}
\newcommand{\RepValDeltaCentroidSeedB}{+1.23}
\newcommand{\RepPlacementDelta}{+0.70}
\newcommand{\RepPlacementDeltaCI}{[+0.30,\,+1.11]}

\newcommand{\RepDeltaEnvelopeSeedA}{+0.92}
\newcommand{\RepDeltaEnvelopeSeedACI}{[+0.39,\,+1.44]}
\newcommand{\RepValDeltaEnvelopeSeedA}{+0.30}
\newcommand{\RepDeltaEnvelopeSeedB}{+0.75}
\newcommand{\RepDeltaEnvelopeSeedBCI}{[+0.27,\,+1.24]}
\newcommand{\RepValDeltaEnvelopeSeedB}{+1.50}
\newcommand{\RepBudgetShift}{+0.26}
\newcommand{\RepBudgetShiftCI}{[-0.15,\,+0.66]}
\newcommand{\RepValBudgetShift}{-0.32}

\newcommand{\RepDeltaMeanCentroid}{+0.96}
\newcommand{\RepDeltaOrigCentroid}{+1.11}

\newcommand{\RepValDeltaOrigCentroid}{+0.93}
\newcommand{\RepDeltaMeanEnvelope}{+0.92}
\newcommand{\RepDeltaOrigEnvelope}{+1.14}

\newcommand{\RepValDeltaOrigEnvelope}{+1.30}

\newcommand{\RepMeanAUCRandom}{0.8624}
\newcommand{\RepMeanAUCCentroid}{0.8721}
\newcommand{\RepMeanAUCEnvelope}{0.8716}

\newcommand{\RepPoolPatchMaxAUCRandom}{0.8483}
\newcommand{\RepPoolPatchMaxAUCCentroid}{0.8625}
\newcommand{\RepPoolPatchMaxAUCEnvelope}{0.8553}
\newcommand{\RepPoolPatchMaxDeltaCentroid}{+1.42}
\newcommand{\RepPoolPatchMaxDeltaEnvelope}{+0.69}
\newcommand{\RepPoolSliceMaxAUCRandom}{0.8449}
\newcommand{\RepPoolSliceMaxAUCCentroid}{0.8637}
\newcommand{\RepPoolSliceMaxAUCEnvelope}{0.8633}
\newcommand{\RepPoolSliceMaxDeltaCentroid}{+1.88}
\newcommand{\RepPoolSliceMaxDeltaEnvelope}{+1.84}
\newcommand{\RepPoolMeanMaxAUCRandom}{0.8553}
\newcommand{\RepPoolMeanMaxAUCCentroid}{0.8709}
\newcommand{\RepPoolMeanMaxAUCEnvelope}{0.8719}
\newcommand{\RepPoolMeanMaxDeltaCentroid}{+1.56}
\newcommand{\RepPoolMeanMaxDeltaEnvelope}{+1.66}
\newcommand{\RepPoolDeltaMinCentroid}{+1.42}
\newcommand{\RepPoolDeltaMaxCentroid}{+1.88}
\newcommand{\RepPoolDeltaMinEnvelope}{+0.69}
\newcommand{\RepPoolDeltaMaxEnvelope}{+1.84}
\newcommand{\RepBudgetURandomCB}{102.2}
\newcommand{\RepBudgetLRandomCB}{159.2}
\newcommand{\RepBudgetDRandomCB}{57.1}
\newcommand{\RepBudgetCRandomCB}{79.5}
\newcommand{\RepCBBatches}{3{,}740}
\newcommand{\RepCBHiddenDiff}{0.00}
\newcommand{\RepBudgetURandom}{113.1}
\newcommand{\RepBudgetLRandom}{160.2}
\newcommand{\RepBudgetDRandom}{47.1}
\newcommand{\RepBudgetCRandom}{70.5}
\newcommand{\RepBudgetUCentroid}{102.5}
\newcommand{\RepBudgetLCentroid}{160.2}
\newcommand{\RepBudgetDCentroid}{57.7}
\newcommand{\RepBudgetCCentroid}{80.8}
\newcommand{\RepPurityCentroid}{46.0}
\newcommand{\RepPurityRandomCB}{36.7}
\newcommand{\RepPurityRandom}{36.0}
\newcommand{\RepPurityViews}{190}
\newcommand{\RepCBMatchImages}{960}
\newcommand{\RepCBMismatches}{0}
\newcommand{\RepMDMAERandomSeedA}{3.55}
\newcommand{\RepMDRMSERandomSeedA}{4.72}
\newcommand{\RepMDSpearmanRandomSeedA}{0.52}
\newcommand{\RepMDMAERandomSeedB}{3.49}
\newcommand{\RepMDRMSERandomSeedB}{4.64}
\newcommand{\RepMDSpearmanRandomSeedB}{0.53}
\newcommand{\RepMDMAECentroidOrig}{3.45}
\newcommand{\RepMDRMSECentroidOrig}{4.57}
\newcommand{\RepMDSpearmanCentroidOrig}{0.53}
\newcommand{\RepMDMAECentroidSeedA}{3.47}
\newcommand{\RepMDRMSECentroidSeedA}{4.60}
\newcommand{\RepMDSpearmanCentroidSeedA}{0.53}
\newcommand{\RepMDMAECentroidSeedB}{3.46}
\newcommand{\RepMDRMSECentroidSeedB}{4.58}
\newcommand{\RepMDSpearmanCentroidSeedB}{0.54}
\newcommand{\RepMDMAEEnvelopeOrig}{3.46}
\newcommand{\RepMDRMSEEnvelopeOrig}{4.60}
\newcommand{\RepMDSpearmanEnvelopeOrig}{0.54}
\newcommand{\RepMDMAEEnvelopeSeedA}{3.50}
\newcommand{\RepMDRMSEEnvelopeSeedA}{4.64}
\newcommand{\RepMDSpearmanEnvelopeSeedA}{0.54}
\newcommand{\RepMDMAEEnvelopeSeedB}{3.44}
\newcommand{\RepMDRMSEEnvelopeSeedB}{4.60}
\newcommand{\RepMDSpearmanEnvelopeSeedB}{0.53}
\newcommand{\RepMDMAERandomCBSeedA}{3.49}
\newcommand{\RepMDRMSERandomCBSeedA}{4.67}
\newcommand{\RepMDSpearmanRandomCBSeedA}{0.52}
\newcommand{\RepMDBaseMAE}{4.65}
\newcommand{\RepMDBaseRMSE}{5.96}
\newcommand{\RepMDN}{1{,}466}
\newcommand{\RepMDMAEMin}{3.44}
\newcommand{\RepMDMAEMax}{3.55}
\newcommand{\RepMDDeltaMAECentroidSeedA}{-0.08}
\newcommand{\RepMDDeltaMAECentroidSeedACI}{[-0.16,\,-0.00]}
\newcommand{\RepMDDeltaSpearmanCentroidSeedA}{+0.01}
\newcommand{\RepMDDeltaMAEEnvelopeSeedA}{-0.05}
\newcommand{\RepMDDeltaMAEEnvelopeSeedACI}{[-0.13,\,+0.03]}
\newcommand{\RepMDDeltaSpearmanEnvelopeSeedA}{+0.02}
\newcommand{\RepMDDeltaMAERandomCBSeedA}{-0.06}
\newcommand{\RepMDDeltaMAERandomCBSeedACI}{[-0.13,\,+0.01]}
\newcommand{\RepMDDeltaSpearmanRandomCBSeedA}{-0.00}
\newcommand{\RepMDDeltaMAECentroidSeedB}{-0.03}
\newcommand{\RepMDDeltaMAECentroidSeedBCI}{[-0.10,\,+0.05]}
\newcommand{\RepMDDeltaSpearmanCentroidSeedB}{+0.01}

\newcommand{\RepNHeadSeedsWord}{five}

\newcommand{\RepResultsSeeds}{Table~\ref{tab:replication} reports the original encoders and two additional completed continuations per strategy at epoch~\RepEndpoint{}. These runs share the epoch-25 checkpoint and are reported descriptively.}
\newcommand{\RepResultsControl}{With these budgets, \textsc{random-cb} differs from \textsc{random} of the same seed by $\RepBudgetShift$ pp. The placement difference, \ArmBest{} minus \textsc{random-cb}, is $\RepPlacementDelta$ pp (descriptive). The comparison is a single continuation per arm, and the two policies differ only modestly in  the proportion of target patches within the retina ($\RepPurityCentroid$\% versus $\RepPurityRandomCB$\%), so a null placement difference would be weak evidence that placement does not matter.}
\newcommand{\RepResultsPooling}{For seed A, the alternative poolings give differences from \textsc{random} ranging from $\RepPoolDeltaMinCentroid$--$\RepPoolDeltaMaxCentroid$ pp for \ArmBest{} and from $\RepPoolDeltaMinEnvelope$--$\RepPoolDeltaMaxEnvelope$ pp for \textsc{envelope} (Table~\ref{tab:pooling}). These comparisons are exploratory.}
\newcommand{\RepDiscussionLead}{Anatomy-guided rectangular target placement improved glaucoma AUC in the original comparisons. Because the strategies also change target size, overlap and visible context, these results do not isolate the effect of target placement.}
\newcommand{\RepLimitations}{The additional experiments include two completed continuations per primary strategy and one matched-budget control. Their limited number and shared checkpoint do not establish a ranking across independent pretraining runs.}

\newcommand{\RepAppendixResults}{Table~\ref{tab:replication_runs} lists every evaluated encoder with its paired interval and Table~\ref{tab:pooling} the pooling variants, and Figure~\ref{fig:seed_spread} shows the spread across runs. In glaucoma cases of the test split ($n=\RepMDN$), ridge regression on the frozen features predicts visual-field mean deviation with a mean absolute error from $\RepMDMAEMin$ to $\RepMDMAEMax$ dB across encoders, against $\RepMDBaseMAE$ dB for the training mean (Table~\ref{tab:md}). Mean deviation partly defines the glaucoma label, so this is a severity endpoint on the same cohort, not an independent task.}

\newcommand{\ArmBest}{\textsc{centroid}}

\title{Where to Predict in Retinal OCT?\\
Anatomy-Guided Target Selection for I-JEPA}

\author{%
  Yunchu Feng \\
  University of Pennsylvania \\
  \texttt{yfeng26@engineering.upenn.edu} \\
  \And
  Wai Tak Lau \\
  Columbia University \\
  \texttt{wl2822@columbia.edu} \\
  \And
  Kaveri A. Thakoor \\
  Columbia University \\
  \texttt{kat2193@columbia.edu} \\
}

\begin{document}

\maketitle

\begin{abstract}
Self-supervised learning  aims to learn meaningful representations from unlabeled data, potentially allowing models to capture subtle anatomical variations across patients and reduce the amount of labeled data needed.    Image-based Joint-Embedding
Predictive Architecture (I-JEPA) learns by predicting target-region
representations from visible context rather than reconstructing pixels.
However, its uniform target placement does not account for the  anatomical structure of retinal OCT, where retinal tissue occupies only part of
each B-scan. We investigate anatomy-guided target selection using
image intensity and predicted retinal segmentation while retaining the
I-JEPA architecture and latent prediction objective.
We compare anatomy-guided rectangular targets, anatomy-shaped targets
and coverage-constrained placement with uniform masking, and evaluate
the learned representations using linear probing for glaucoma classification
on FairVision.   
 We found that rectangular targets guided by image intensity achieved the highest linear-probe AUC in our comparisons. We show that anatomically meaningful target selection is promising for retinal OCT representation learning, while greater anatomical specificity did not consistently improve performance.
\end{abstract}

\begin{figure}[t]
\centering
\includegraphics[width=\linewidth]{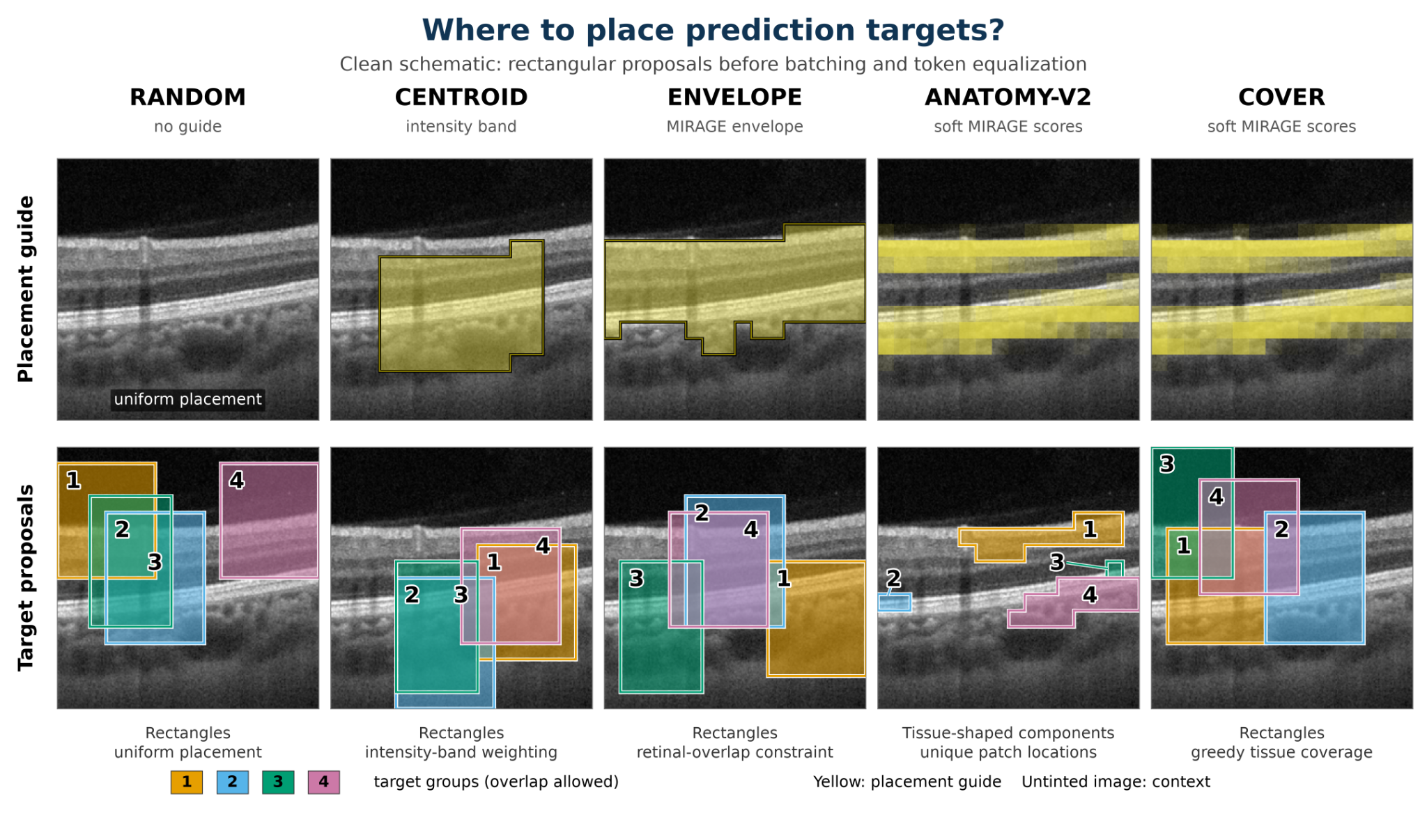}
\caption{  Top: placement guides. Bottom: target  masks. Colours and numbers identify target groups; overlaps are allowed. Illustrative B-scan adapted from Kulyabin et al., OCTDL (arXiv:2312.08255v3), under \href{https://creativecommons.org/licenses/by/4.0/}{CC BY 4.0}; overlays added. }
\label{fig:target_selection}
\end{figure}

\section{Introduction}

Optical coherence tomography (OCT)  visualizes retinal layers
and supports glaucoma assessment~\citep{maetschke2019glaucoma,ran2019glaucoma,noury2022glaucoma}.
Self-supervised pretraining  is widely used in medical imaging foundation models to  learn transferable representations and improve downstream performance.
 ~\citep{azizi2021big,zhou2023retfound,qiu2024visionfm}.
 Image-based Joint-Embedding Predictive Architecture (I-JEPA) in particular has gained interest with extension to both robotics and medical images ~\citep{Judkiewicz2026-me}, motivated by learning more semantically meaningful features by predicting target representations instead of pixels.   
The spatial organization of OCT makes the choice of pretraining task
particularly relevant: retinal tissue forms a relatively narrow region within
a B-scan, surrounded by  the vitreous (the transparent gel anterior to the retina) and other low-intensity background regions. Our interest in anatomical guidance also arose from exploratory
feature-occlusion maps of glaucoma classifiers initialized from
uniform-masking I-JEPA and then fine-tuned with diagnosis labels, in which
selected maps suggested prediction sensitivity near retinal structures
(Appendix~\ref{app:interp}).

Standard I-JEPA places rectangular targets uniformly across the image.
In retinal OCT, many such targets include little retinal tissue. This raises
our central question: \emph{can I-JEPA benefit from selecting prediction
targets according to retinal anatomy rather than uniformly across the image?}
Figure~\ref{fig:target_selection} illustrates the change in target placement.
Semantic and anatomy-guided masked modelling provide positive motivation
for this question~\citep{li2022semmae,kakogeorgiou2022attmask,li2024anatomask,wang2025maskwhatmatters}.
However, selecting a target also removes information from the context.
The useful learning task depends on both what is predicted and what remains
visible.

We investigate this question through five masking families: uniform
\textsc{random}, the segmentation-free \ArmBest{}, segmentation-guided
\textsc{envelope}, anatomy-shaped targets (two versions) and coverage-constrained
\textsc{cover}, the last three guided by MIRAGE retinal
segmentation~\citep{morano2025mirage}. They vary target location, shape or
coverage within the unchanged I-JEPA architecture and latent objective.
We evaluate the resulting B-scan representations
by pooling them across an OCT volume and fitting a linear glaucoma classifier. We first pretrain our model for 25 epochs as initialization. Then we apply different masking strategies for the rest of pretraining, and evaluate glaucoma classification performance. We found that simpler methods based on image intensity outperform masking methods using more detailed anatomical guidance.

\paragraph{Our contributions are as follows:}
\setlength{\itemsep}{0pt}\setlength{\parskip}{0pt}
\begin{itemize}
\setlength{\itemsep}{0pt}\setlength{\parskip}{0pt}
\item An empirical comparison of intensity-guided, segmentation-guided and
      uniform target selection for I-JEPA on retinal OCT.
\item Linear-probe results showing higher glaucoma classification AUC with anatomy-guided rectangular targets in the original runs, while more anatomically specific strategies did not consistently improve performance.

\end{itemize}

\section{Related work}

\paragraph{Masked image modelling and JEPAs.}

Masked image modelling learns visual representations by predicting hidden content from visible context.
Reconstruction-based approaches define this prediction in terms of pixels or discrete visual tokens~\citep{he2022mae,xie2022simmim,bao2022beit}.
A complementary line of work uses teacher outputs or latent representations as prediction targets, moving the learning task into representation space~\citep{zhou2022ibot,baevski2022data2vec,assran2023ijepa}.
Masked prediction has also been explored in video and volumetric data~\citep{bardes2024vjepa,chen2023mim3d,taleb20203dssl}, while JEPA variants investigate different objectives and prediction structures~\citep{balestriero2025lejepa,he2025dseqjepa}.
Alongside these choices, target selection determines which image regions the model learns to predict.
Our study focuses on this spatial aspect in retinal OCT, retaining the I-JEPA objective while guiding target selection using retinal anatomy.

\paragraph{Region selection and prediction structure.}
DSeq-JEPA combines attention-derived region selection with sequential
prediction~\citep{he2025dseqjepa}. Its highest-ranked region supplies
context for subsequent predictions, and its ablation favours the joint design
over region selection alone. We instead use retinal anatomy with parallel
I-JEPA prediction, where the guidance map, selected targets and available
context jointly define the learning task. I-JEPA's own masking analysis emphasizes both
sufficiently large targets and informative context~\citep{assran2023ijepa}.

\paragraph{Informed masking.}

Mask selection can shape self-supervision by emphasizing particular image regions.
Existing approaches derive masks from learned policies~\citep{shi2022adios}, semantic or attention-based guidance~\citep{li2022semmae,kakogeorgiou2022attmask}, or prediction difficulty~\citep{wang2023hpm,chen2023automae}.
In medical imaging, anatomical structure also guides masking~\citep{li2024anatomask,ceballosarroyo2026anatomymae,huang2026vasomim}.
These approaches motivate anatomy-guided target selection, although their prediction objectives, guidance sources and context construction differ from those studied here.

\paragraph{Medical and retinal self-supervised learning.}

Medical pretraining can incorporate modality-specific structure into region selection.
US-JEPA uses a frozen ultrasound teacher, with a region-conditioned variant restricting context and target sampling to valid ultrasound content~\citep{radhachandran2026usjepa}.
In retinal imaging, foundation models pretrain on large retinal corpora~\citep{zhou2023retfound,qiu2024visionfm,liu2026octcube}, while OCT-specific work explores masked image modelling~\citep{pissas2024octmim}.
We use image intensity and MIRAGE retinal segmentation~\citep{morano2025mirage} to guide target selection within the EMA-based I-JEPA framework, retaining its architecture and parallel prediction structure.

\section{Background and target-selection framework}
\label{sec:policies}

We first define the I-JEPA prediction task, then describe how each masking
strategy selects its targets. The architecture and latent prediction loss
are shared; differences in guide sources, target size and final context
are part of the strategy comparisons.

\begin{figure}[t]
\centering
\includegraphics[width=\linewidth]{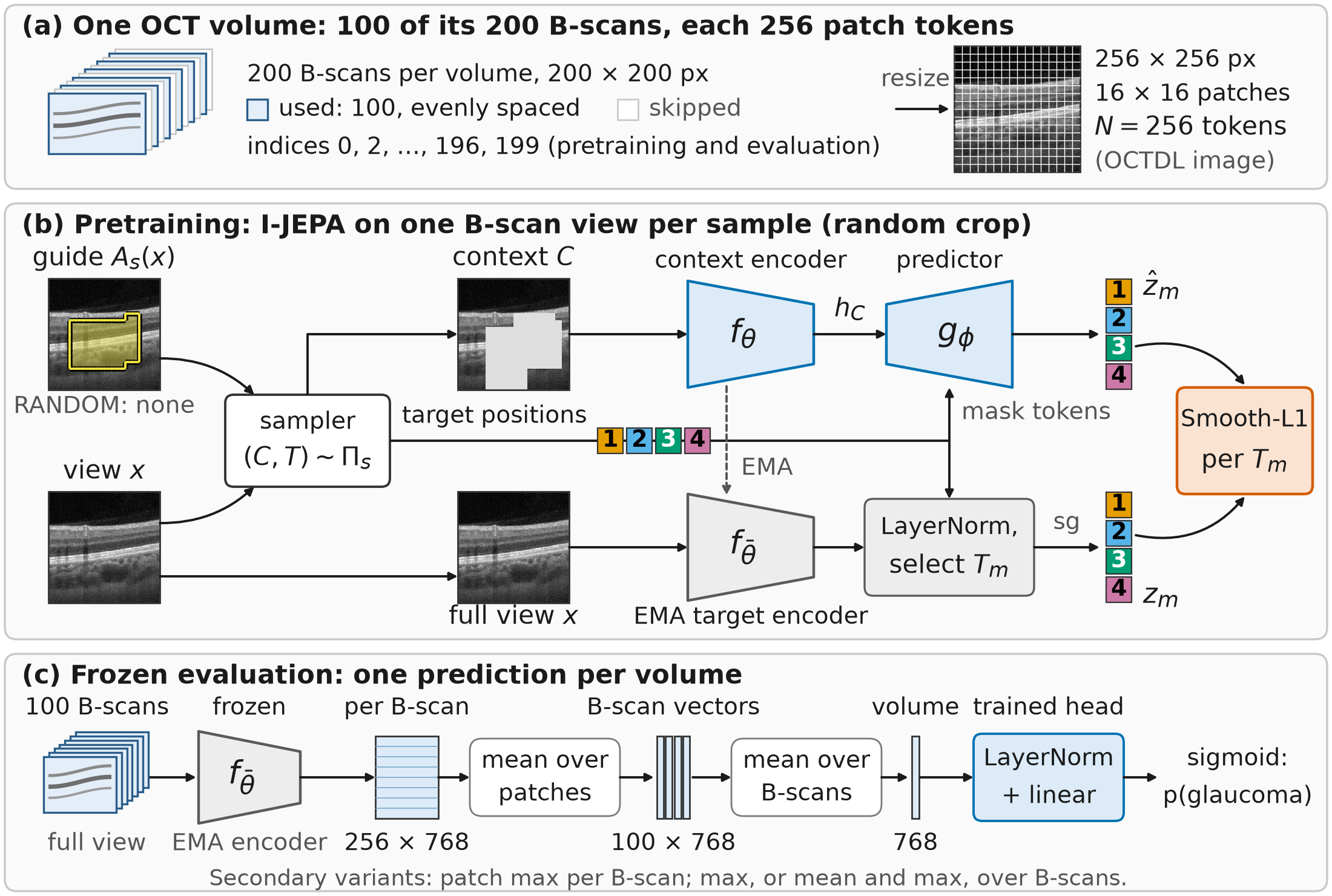}
    \caption{Overview of OCT pretraining and glaucoma classification. (a) Each OCT volume contains 200 two-dimensional slices (B-scans); we select 100 evenly spaced slices. (b) For each slice, the context encoder processes visible patches, and a predictor estimates the features of masked target regions. Standard I-JEPA uses randomly placed rectangles; our strategies change target selection. A Smooth-L1 loss compares these predictions with features from a target encoder that sees the full slice. The loss trains the context encoder and predictor, while the target encoder follows a moving average of the context encoder's weights. (c) For evaluation, the trained target encoder is frozen. Features are averaged across patches and slices, and a linear classifier predicts glaucoma for the volume. The illustrative B-scan is sourced as in Figure~\ref{fig:target_selection}.}
\label{fig:policies}
\end{figure}

\subsection{I-JEPA preliminaries}

Let $x$ be a B-scan partitioned into $N$ patches, and let $C$ denote the
indices of the patches supplied to the context encoder. Target group $T_m$
contains $K_m$ patch indices, with $m=1,\ldots,M$. We treat $T_m$ as an
index list because some implementations repeat indices when equalizing target
lengths. The context excludes the union of target positions.
Writing $p_C$ and $p_{T_m}$ for the corresponding positional embeddings,
the context and prediction paths are
\begin{equation}
 h_C=f_\theta(x;C), \qquad
 \widehat z_m=g_\phi(h_C,p_C,p_{T_m}).
 \label{eq:prediction}
\end{equation}
Here $f_\theta(x;C)$ embeds only the selected patches at their original image
positions. The predictor projects $h_C$ to its own feature dimension and
adds its own positional embeddings $p_C$; target queries combine a learned
mask token with $p_{T_m}$. The target encoder sees the full B-scan. Its output is normalized
per patch before selecting the target positions:
\begin{equation}
 z_m=\operatorname{sg}\!\left(
       [\operatorname{LN}(f_{\bar\theta}(x))]_{T_m}
      \right), \qquad
 \bar\theta\leftarrow \mu\bar\theta+(1-\mu)\theta .
 \label{eq:teacher}
\end{equation}
The operator $\operatorname{sg}$ stops gradients, $\operatorname{LN}$ is
LayerNorm across feature dimensions, and $\mu$ is the EMA coefficient.
The predictor uses the same context for each target group; predictions are
not conditioned on earlier target predictions.

Each target and prediction is a $K_m\times D$ matrix.
For feature dimension $D$, the implementation minimizes the mean Smooth-L1
loss over target tokens and feature coordinates:
\begin{equation}
 \mathcal L =
 \frac{1}{D\sum_{m=1}^{M}K_m}
 \sum_{m=1}^{M}\sum_{k=1}^{K_m}\sum_{d=1}^{D}
 \rho\!\left(\widehat z_{m,k,d}-z_{m,k,d}\right).
 \label{eq:loss}
\end{equation}
The Smooth-L1 function $\rho$ uses the standard unit transition between its
quadratic and linear regions. We call the $\sum_m K_m$ supervised target tokens
\emph{loss slots}; repeated or overlapping indices add loss slots, and the mean
is taken over all loss slots of a batch (Appendix~\ref{app:covbug}).
Figure~\ref{fig:policies} shows the pretraining and evaluation paths.

\subsection{Guided target selection}

Background patches above and below the retina often have low intensity
variation, whereas  retinal patches carry the layered structure. Guided strategies
direct targets toward  retinal structures so that prediction depends more on  them. However, even a visually uniform background patch receives a positional embedding, and its latent representation can incorporate tissue information through attention. Such patches may therefore still provide useful prediction targets or visible context. For strategy $s$, let $A_s(x)$ denote its patch-grid guidance:
an intensity-derived band for \ArmBest{}, a repaired segmentation envelope
for \textsc{envelope}, or the InnerRetina and choroid soft score maps for the anatomy-shaped
and coverage strategies. \ArmBest{} is an intensity-guided  retinal localization
baseline, not a segmentation model. We write the joint masking procedure as
\begin{equation}
 (C,T)\sim\Pi_s(\,\cdot\mid A_s(x)),
 \qquad T=(T_1,\ldots,T_M).
 \label{eq:selection}
\end{equation}
$\Pi_s$ includes target placement and context construction,
including batch-dependent token equalization.
A candidate context block is sampled, target positions are
excluded, and token lists are shortened as needed for batching. Only the
remaining context tokens enter $f_\theta$. Consequently, tissue outside the
target union need not be visible to the encoder.
We report \emph{unique targets} (cells in the target union), \emph{tissue
hidden} (share of tissue cells inside that union), \emph{purity} (share of
target cells on tissue) and \emph{context} (share of the grid supplied to the
context encoder); the target union and the context need not be complementary.

\subsection{Compared masking strategies}

All policies request $M{=}4$ target blocks and one context block
(Table~\ref{tab:strategies}, Figure~\ref{fig:target_selection}). Rectangle
strategies share nominal scale and aspect-ratio ranges, but placement and
collation can change unique targets and encoder-visible context. We denote the segmentation scores by $\widehat Y=s_\psi(x)$, with guide-specific class definitions (Section~\ref{sec:setup} and Appendix~\ref{app:repro}). We follow ~\citep{assran2023ijepa} for  rectangular mask sizes  and ratio except for  \textsc{anatomy}.  

\begin{description}
\setlength{\itemsep}{1pt}\setlength{\parskip}{0pt}
\item[\textsc{random} (baseline).]   Masking strategies adopted from I-JEPA ~\citep{assran2023ijepa}, where four rectangular masks are randomly placed and no image content is used for masking. 

\item[\textsc{envelope} (random-within-retina).] 

{We sample the same number of rectangles, using the same size and width-to-height ratio distributions as \textsc{random}. We then place them randomly where they overlap sufficiently with a repaired retinal envelope estimated by MIRAGE~\mbox{\citep{morano2025mirage}}. The guide uses MIRAGE-Large fine-tuned for layer segmentation on GOALS circumpapillary B-scans~\mbox{\citep{fang2022goals}}. Its foreground Dice score of $\CRMirageDice$ on held-out GOALS images describes performance on GOALS, not FairVision, which lacks layer annotations and uses a different scan protocol. We combine the predicted inner-retinal and choroid regions and fill the unlabeled middle-retina gap in the original predicted label map, then transform the guide and image together.}

\item[\ArmBest{} (segmentation-free).] To account for variation in the vertical position of retinal tissue across FairVision B-scans, we first estimate the band center.
Let $\bar x_{ij}$ be the mean intensity
  of the patch in row $i$ and column $j$, minus the smallest patch mean in the
  image. For columns with $\sum_i\bar x_{ij}>\varepsilon$, where $\varepsilon=10^{-6}$, the band
  center in column $j$ is the intensity-weighted row centroid
  \begin{equation}
   c_j=\textstyle\sum_i i\,\bar x_{ij}\big/\sum_i \bar x_{ij}.
   \label{eq:centroid}
  \end{equation}
  Columns with negligible total weight use the image-wide intensity centroid, falling back to the vertical grid center when the image-wide weight is also negligible. We denote the centers averaged over a three-column neighborhood by $\tilde c_j$.
  A band of fixed height centered on $\tilde c_j$
  across the central columns (Appendix~\ref{app:repro}) weights the placement of
  four rectangular targets; it is not itself the target shape. No segmentation
  model or annotation of any kind is used.

\item[\textsc{anatomy-v1} / \textsc{anatomy-v2}.] To align target shape with retinal anatomy, we construct target regions from the predicted retinal-class score maps.
Connected components
  are grown on soft MIRAGE retinal-class scores from a frozen feature adapter
  (Appendix~\ref{app:repro}), so target \emph{shape} follows  retinal anatomy. Each
  component is resampled to exactly $K{=}16$ indices, subsampled when larger and
  topped up with repeated indices when smaller. v2 also bridges diagonal adjacencies.

\item[\textsc{cover} $r$.] This strategy aims to balance anatomical target coverage with the retention of  retinal information in the visible context.
Targets are chosen greedily to cover anatomy while
  leaving a fraction $r$ of supported soft guide mass outside the target union. The trained configuration uses
  $r{=}0.21$. This condition does not guarantee the same fraction survives
  final encoder-context collation, and post-placement target truncation changes
  the intended coverage (Appendix~\ref{app:covbug}).
\end{description}

\begin{table}[t]
\centering
\caption{Compared strategies. Variable-length rectangle targets are shortened
to the shortest target in the batch (Appendix~\ref{app:covbug}). Delivered
target and context budgets are reported in Table~\ref{tab:geometry}.}
\label{tab:strategies}
\footnotesize
\setlength{\tabcolsep}{4pt}
\begin{tabular}{lllll}
\toprule
policy & guide $A_s(x)$ & target shape & placement & target length \\
\midrule
\textsc{random} & none & rectangles & uniform & variable \\
\textsc{envelope} & hard MIRAGE envelope & rectangles & rejection-sampled on envelope & variable \\
\ArmBest{} & intensity-centroid band & rectangles & weighted by band overlap & variable \\
\textsc{anatomy-v1/v2} & soft MIRAGE scores & connected components & grown on scores & fixed ($K$) \\
\textsc{cover} & soft MIRAGE scores & rectangles & greedy coverage & variable \\
\bottomrule
\end{tabular}
\end{table}

\section{Experimental setup}
\label{sec:setup}

\paragraph{Data.}
We use FairVision OCT volumes with binary glaucoma
labels~\citep{luo2024fairvision} and the fixed released splits.
Training and validation contain 6{,}000 and 1{,}000 volumes, respectively.
Each volume contributes 100 of its 200 B-scans, evenly spaced and resized to
$256{\times}256$ (Figure~\ref{fig:policies}a).
The test split contains \Ntest{} volumes (\Npos{} with glaucoma).
Diagnosis labels are used only for downstream evaluation, not JEPA pretraining.
Demographic attributes are not model inputs.

\paragraph{Pretraining.} Patch-level I-JEPA, ViT-Base, $256{\times}256$ crops,
patch size 16, predictor depth 6, EMA target encoder. The original
continuations fork from a shared epoch-25 checkpoint (linear-probe AUC $\AUCAncestor$) of uniform-masking
pretraining, with matched nominal optimizer, learning-rate and
weight-decay schedules and effective batch size (512).
The probability of guided placement rises linearly to full strength between
epochs 25 and 30. The lineage of the early \textsc{anatomy-v1} continuation is
described in Appendix~\ref{app:excluded}.
The encoder processes B-scans independently, without attention across
B-scans. Table~\ref{tab:strategies} and Appendix~\ref{app:repro} list the
remaining differences between strategies.

\textsc{random}, \ArmBest{}, \textsc{envelope} and \textsc{cover} reach
epoch 100. \textsc{anatomy-v2} reaches epoch 75; its epoch-100 result is
unmeasured. We compare
strategies at matched pretraining epochs, with epoch 50 as the common
endpoint for the five main strategies. Excluded continuations are listed
separately in Appendix~\ref{app:excluded}.

\textbf{Segmentation guides.} \textsc{envelope} uses a GOALS-tuned MIRAGE-Large model with a ConvNeXt decoder and four classes: Elsewhere (other retinal tissue and image background), retinal nerve fiber layer (RNFL), ganglion cell and inner plexiform layers (GCIPL), and choroid. Its foreground predictions are combined and repaired into a retinal envelope; its held-out GOALS foreground Dice is $0.925$. The MIRAGE-Base soft guides used by \textsc{anatomy} and \textsc{cover} combine RNFL and GCIPL as InnerRetina, retaining choroid separately. This model also has Elsewhere and ignored void output channels; only the InnerRetina and choroid soft score maps are retained for masking. Appendix~\ref{app:repro} details the label harmonization and adapters.

\paragraph{Additional continuations and controls.} The main comparison uses one continuation per strategy. We also report two additional completed continuations each of \textsc{random}, \ArmBest{} and \textsc{envelope} from the same epoch-25 checkpoint, plus one uniform-placement control matched to \ArmBest{}'s per-image target and context budgets. Seeds vary data order, crops and mask draws. These additional comparisons are descriptive and do not represent independent initializations. Checkpoint decisions used validation AUC. Details are in Appendix~\ref{app:replication}.

\paragraph{Evaluation.} The EMA target encoder is frozen and encodes each B-scan
without cropping. We average patch features within each
B-scan, then average across the volume's B-scans and train a LayerNorm--linear
head (Figure~\ref{fig:policies}c).
Each head uses a 50-epoch budget and is selected by validation AUC.
We report test AUC with percentile bootstrap intervals and paired differences
(\Nboot{} class-stratified resamples, fitted heads fixed), together with
DeLong tests~\citep{delong1988roc}. These intervals describe test-case
sampling, not variation across pretraining runs.
AUC differences and their confidence intervals are reported in percentage points (pp), unless otherwise stated. The original \textsc{random}, \ArmBest{} and \textsc{envelope} probes used fp16 mixed precision because their evaluation configurations retained the default automatic mixed-precision setting; \textsc{anatomy-v2} and \textsc{cover} explicitly disabled it and used fp32. In Table~\ref{tab:main}, each difference uses a \textsc{random} baseline matched in epoch and probe precision. Repeating linear probing with the same frozen encoders in fp32 changed AUC by at most $2\times10^{-4}$ (0.02 pp). Further precision-control results are
described in Appendix~\ref{app:fp32}.

\section{Results and analysis}
\label{sec:results}

These are strategy comparisons, not factorial ablations: each original strategy
has one pretraining continuation and may change several aspects of the
prediction task. Section~\ref{sec:replication} reports additional completed continuations.

\subsection{Comparison of target-selection strategies}

\begin{table}[t]
\centering
\caption{Mean-pooled linear-probe test AUC on FairVision (upper block: fp16; lower block: fp32). $\Delta$ is the AUC difference from \textsc{random} (pp); only positive differences are bold.}
\label{tab:main}
\footnotesize
\setlength{\tabcolsep}{4pt}
\begin{tabular}{llcccccc}
\toprule
& & \multicolumn{2}{c}{epoch 50} & \multicolumn{2}{c}{epoch 75} & \multicolumn{2}{c}{epoch 100} \\
\cmidrule(lr){3-4}\cmidrule(lr){5-6}\cmidrule(lr){7-8}
strategy & guidance signal & AUC & $\Delta$ (pp) & AUC & $\Delta$ (pp) & AUC & $\Delta$ (pp) \\
\midrule
\textsc{random}   & none (null)                      & \AUCRandomEpFifty  & ---                              & \AUCRandomEpSeventyFive  & ---                              & \AUCRandomEpHundred  & ---  \\
\textsc{envelope} & segmenter (location only)        & \AUCEnvelopeEpFifty & $\mathbf{\DEnvelopeRandomEpFifty}$ & \AUCEnvelopeEpSeventyFive & $\mathbf{\DEnvelopeRandomEpSeventyFive}$ & \AUCEnvelopeEpHundred & $\mathbf{\DEnvelopeRandomEpHundred}$ \\
\ArmBest{}   & intensity centroid, no segmenter & \AUCOracleEpFifty   & $\mathbf{\DOracleRandomEpFifty}$   & \AUCOracleEpSeventyFive   & $\mathbf{\DOracleRandomEpSeventyFive}$   & \AUCOracleEpHundred & $\mathbf{\DOracleRandomEpHundred}$ \\
\midrule
\textsc{anatomy-v2} & segmenter (target shape) & \TAUCAnatomyTwoEpFifty & $\DAnatomyTwoRandomEpFifty$ & \TAUCAnatomyTwoEpSeventyFive & \TAnatomyTwoRandomEpSeventyFive & \TAUCAnatomyTwoEpHundred & \TAnatomyTwoRandomEpHundred \\
\textsc{cover} $r{=}0.21$ & segmenter (coverage) & \TAUCCoverEpFifty & $\DCoverRandomEpFifty$ & \TAUCCoverEpSeventyFive & \TCoverRandomEpSeventyFive & \TAUCCoverEpHundred & \TCoverRandomEpHundred \\
\midrule
\multicolumn{2}{l}{reported: supervised 3D ResNet~\citep{luo2024fairvision}} & \multicolumn{6}{l}{\CRFairVisionAUC{} (trained end to end on all B-scans; not this probe)} \\
\bottomrule
\end{tabular}
\end{table}

\paragraph{Anatomy-guided rectangles.}
At the common endpoint, epoch 50, \textsc{envelope} and the segmentation-free
\ArmBest{} exceed uniform masking by $\DEnvelopeRandomEpFifty$ pp (CI
$\DEnvelopeRandomEpFiftyCI$) and $\DOracleRandomEpFifty$ pp (CI
$\DOracleRandomEpFiftyCI$). Across the three evaluated epochs, gains ranged from 0.62--1.20 pp for \textsc{envelope} and 0.99--1.13 pp for \ArmBest{}
(Figure~\ref{fig:traj}; multiple-comparison adjustment in Appendix~\ref{app:contrasts}).
These gains favour  anatomy-guided target placement in the original runs.
Both strategies also leave more of the grid visible to the context encoder
than \textsc{random} (Table~\ref{tab:geometry}), so these comparisons do not
separate target location from visible context. The gains do not require
targets to contain only retinal tissue: the large rectangles also include
background.

\begin{figure}[t]
\centering
\includegraphics[width=\linewidth]{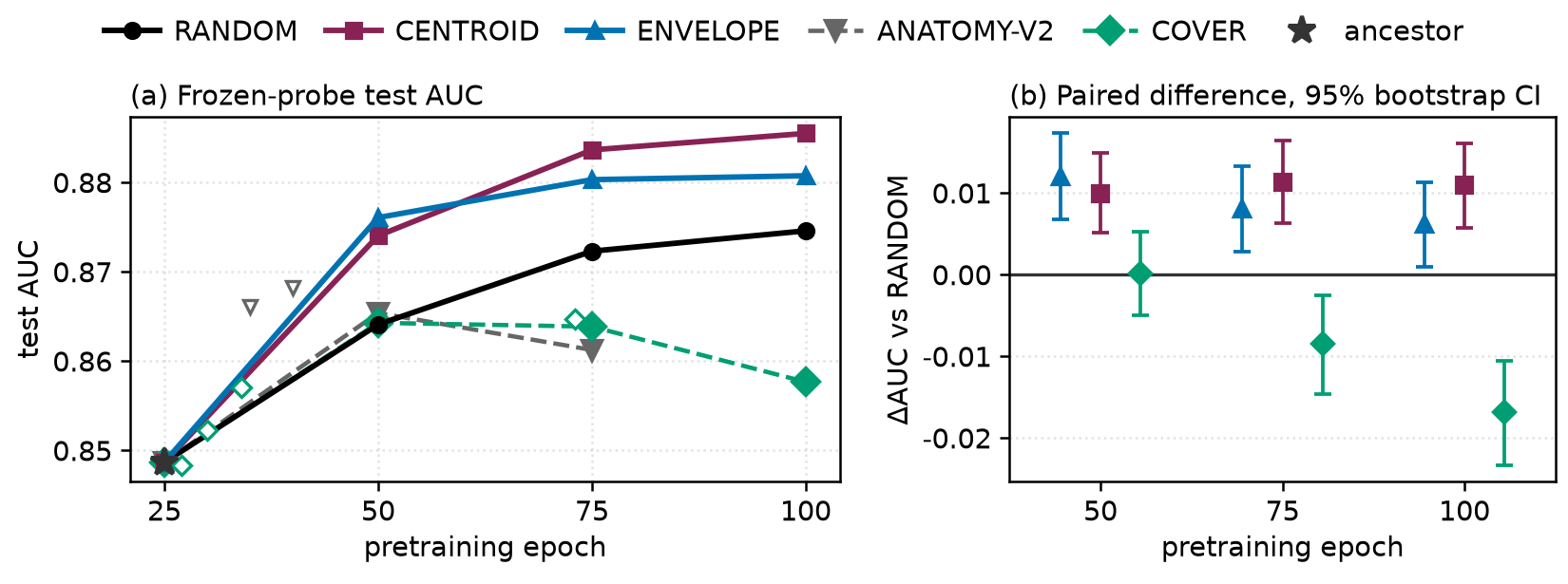}
\caption{ Original-run  linear probe performance on FairVision. (a) Test AUC across pretraining epochs (truncated vertical axis). The star marks the shared epoch-25 checkpoint. Solid and dashed lines denote fp16 and fp32 probes. (b) Paired AUC differences (on the original AUC scale) from precision-matched \textsc{random}, with \Nboot{}-resample percentile bootstrap intervals. Intervals reflect test-case sampling, not variation across pretraining runs.}
\label{fig:traj}
\end{figure}

\paragraph{Anatomy-shaped targets.}
At the matched epoch 50, \textsc{anatomy-v2}
has lower AUC than \textsc{envelope} ($\Delta=\DAnatomyTwoEnvelopeEpFifty$ pp; CI
$\DAnatomyTwoEnvelopeEpFiftyCI$). At epoch 50,
the intervals for both \textsc{anatomy-v2} and \textsc{cover} versus
\textsc{random} include zero.

\paragraph{Coverage-constrained placement.}
\textsc{cover} is close to \textsc{random} at epoch 50 ($\DCoverRandomEpFifty$ pp, CI
$\DCoverRandomEpFiftyCI$) but falls behind by epoch 100
($\DCoverRandomEpHundred$ pp, CI $\DCoverRandomEpHundredCI$), as its own AUC
declines after epoch $\CoverPeakEpoch$ while \textsc{random} keeps improving
(Figure~\ref{fig:traj}). Batching shortens its
targets after placement, so the trained policy is not a controlled
higher-coverage counterpart to \textsc{envelope} (Appendix~\ref{app:covbug}).

In the original single runs, the \ArmBest{}--\textsc{envelope} difference at the final evaluated epoch is $\DOracleEnvelopeEpHundred$ pp (CI $\DOracleEnvelopeEpHundredCI$). The paired \ArmBest{}--\textsc{envelope}
difference intervals include zero at epochs 50 and 75.
A simple intensity-based guide is therefore a useful comparator for
segmentation-guided methods.

\subsection{Analysis of target and context masks}
\label{sec:results-h3}

\begin{table}[t]
\centering
\caption{Masks delivered at the training batch size of $\CRAuditBatch$,
measured by replaying the sampler code in each strategy's trained configuration
on $\CRAuditViews$ fixed training views. The masks used during the original
training were not recorded. Values are means over $\CRAuditSeedCount$ sampler
seeds of $\CRAuditDraws$ batches each, with crops and block sizes paired across
strategies. The context column also gives the across-seed s.d.
Tissue is a thresholded MIRAGE-derived proxy, not ground-truth annotation.
Column definitions are given in
Appendix~\ref{app:geomprov}. AUC is at the matched epoch 50 (fp32 probes for
\textsc{cover} and \textsc{anatomy-v2}).}
\label{tab:geometry}
\footnotesize
\setlength{\tabcolsep}{4pt}
\begin{tabular}{lcccccc}
\toprule
strategy & tissue hidden & purity & unique targets & loss slots & context (\% of grid) & AUC (ep.~50) \\
\midrule
\textsc{random}     & \CRHidRandom\%     & \CRPurRandom\%     & \CRUniqRandom     & \CRSlotsRandom     & $\CRCtxRandom\pm\CRCtxSDRandom$         & \AUCRandomEpFifty \\
\ArmBest{}          & \CRHidCentroid\%   & \CRPurCentroid\%   & \CRUniqCentroid   & \CRSlotsCentroid   & $\CRCtxCentroid\pm\CRCtxSDCentroid$     & \AUCOracleEpFifty \\
\textsc{envelope}   & \CRHidEnvelope\%   & \CRPurEnvelope\%   & \CRUniqEnvelope   & \CRSlotsEnvelope   & $\CRCtxEnvelope\pm\CRCtxSDEnvelope$     & \textbf{\AUCEnvelopeEpFifty} \\
\textsc{cover}      & \CRHidCover\%      & \CRPurCover\%      & \CRUniqCover      & \CRSlotsCover      & $\CRCtxCover\pm\CRCtxSDCover$           & \AUCCoverEpFifty \\
\textsc{anatomy-v2} & \CRHidAnatomyTwo\% & \textbf{\CRPurAnatomyTwo\%} & \CRUniqAnatomyTwo & \CRSlotsAnatomyTwo & $\CRCtxAnatomyTwo\pm\CRCtxSDAnatomyTwo$ & \AUCAnatomyTwoEpFifty \\
\bottomrule
\end{tabular}
\end{table}

Table~\ref{tab:geometry} reports the masks that reach the model. At the
training batch size, \ArmBest{} leaves $\CRCtxGapCentroid$ points more of the
grid visible than \textsc{random} (pooled 95\% interval over batches
\CRCtxGapCentroidCI), and \textsc{envelope} $\CRCtxGapEnvelope$ points more
(\CRCtxGapEnvelopeCI). With loss slots equal across the rectangle strategies,
\ArmBest{}'s rectangles overlap more and cover fewer unique targets than
\textsc{random}'s (Figure~\ref{fig:target_selection}). \textsc{envelope} differs least from
\textsc{random} in target and context budgets but has the  largest proportion of target patches within the retina
among the rectangle strategies. Appendix~\ref{app:geomprov} gives the measurement protocol.

\textsc{anatomy-v2} has  a high proportion of target patches within the retina but a smaller target union and
more total context than the rectangle strategies.
Its fixed $K{=}16$ indices per target give $4\times16=64$ loss slots,
compared with $\CRSlotsRandom$ for each rectangle strategy, an
explicit target-length choice rather than a property of irregular shapes.
Appendix~\ref{app:geomprov} gives the measurement protocol and the per-image
geometry before batching, and Appendix~\ref{app:delivered_audit} audits the
coverage strategy in complete training-sized batches.

\subsection{Additional continuations and matched-budget control}
\label{sec:replication}

\begin{table}[t]
\centering
\caption{Completed continuations at epoch~\RepEndpoint{}: test AUC averaged over \RepNHeadSeedsWord{} probe-head seeds (fp32; head-seed s.d.\ at most $\RepHeadSDMax$). Original encoders are re-probed with the same protocol. $\Delta$ (pp) is the difference from \textsc{random} of the same continuation. Comparisons are descriptive; intervals are in Table~\ref{tab:replication_runs}.}
\label{tab:replication}
\footnotesize
\setlength{\tabcolsep}{4pt}
\begin{tabular}{lccccc}
\toprule
 & \textsc{random} & \ArmBest{} & \textsc{envelope} & $\Delta$ (pp) \ArmBest{} & $\Delta$ (pp) \textsc{envelope} \\
\midrule
original (re-probed) & \RepAUCRandomOrig & \RepAUCCentroidOrig & \RepAUCEnvelopeOrig & $\RepDeltaOrigCentroid$ & $\RepDeltaOrigEnvelope$ \\
seed A & \RepAUCRandomSeedA & \RepAUCCentroidSeedA & \RepAUCEnvelopeSeedA & $\RepDeltaCentroidSeedA$ & $\RepDeltaEnvelopeSeedA$ \\
seed B & \RepAUCRandomSeedB & \RepAUCCentroidSeedB & \RepAUCEnvelopeSeedB & $\RepDeltaCentroidSeedB$ & $\RepDeltaEnvelopeSeedB$ \\
\bottomrule
\end{tabular}
\end{table}

\begin{table}[t]
\centering
\caption{Matched-budget control at epoch~\RepEndpoint{}, seed A, one continuation per arm. $\Delta$ (pp) is the paired difference from \textsc{random} with a test-case interval. Budgets are per image at the training batch size: unique targets (U), loss slots (L), duplicates (D) and context tokens (C). \textsc{random} and \ArmBest{} come from the audit of Table~\ref{tab:geometry}, \textsc{random-cb} comes from its run's mask statistics ($\RepCBBatches$ batches, mean hidden-area residual $\RepCBHiddenDiff$) and \textsc{random-cb} equals its \ArmBest{} shadow per image ($\RepCBMismatches$ mismatches in $\RepCBMatchImages$ image draws). Purity uses a hard MIRAGE-envelope tissue proxy ($\RepPurityViews$ views), not comparable with Table~\ref{tab:geometry}.}
\label{tab:control}
\scriptsize
\setlength{\tabcolsep}{3pt}
\begin{tabular}{lccccccc}
\toprule
policy & test AUC & $\Delta$ (pp) (interval) & U & L & D & C & purity (\%) \\
\midrule
\textsc{random} & \RepAUCRandomSeedA & --- & \RepBudgetURandom & \RepBudgetLRandom & \RepBudgetDRandom & \RepBudgetCRandom & \RepPurityRandom \\
\textsc{random-cb} & \RepAUCRandomCBSeedA & $\RepBudgetShift$ $\RepBudgetShiftCI$ & \RepBudgetURandomCB & \RepBudgetLRandomCB & \RepBudgetDRandomCB & \RepBudgetCRandomCB & \RepPurityRandomCB \\
\ArmBest{} & \RepAUCCentroidSeedA & $\RepDeltaCentroidSeedA$ $\RepDeltaCentroidSeedACI$ & \RepBudgetUCentroid & \RepBudgetLCentroid & \RepBudgetDCentroid & \RepBudgetCCentroid & \RepPurityCentroid \\
\midrule
\multicolumn{2}{l}{\ArmBest{} $-$ \textsc{random-cb}} & $\RepPlacementDelta$ $\RepPlacementDeltaCI$ & \multicolumn{5}{l}{placement at matched budgets} \\
\bottomrule
\end{tabular}
\end{table}

\paragraph{Repeated continuations.} \RepResultsSeeds{}

\paragraph{Matched-budget control.} \textsc{random-cb} receives \ArmBest{}'s
per-image unique targets, loss slots and visible context but places
its targets uniformly. Its difference from \ArmBest{} therefore reflects
placement at equal budgets, and its difference from \textsc{random} reflects the
change of budget. \textsc{envelope} is not budget-matched in this design.
\RepResultsControl{}

\paragraph{Pooling.} We compare three alternative pooling schemes on seed A; the patch maximum within B-scans may retain localized responses that the patch mean attenuates. \RepResultsPooling{}

\section{Discussion and limitations}
\label{sec:limits}

\RepDiscussionLead{}
Masking studies should report the budgets that reach the encoder at the
training batch size, not only the requested ones (Section~\ref{sec:results-h3}).

\paragraph{Experimental limits.}
Each original strategy has one pretraining continuation, so its test-case
confidence intervals do not establish a ranking over
retrainings~\citep{bouthillier2021variance}. The additional experiments use
continuations from the same epoch-25 checkpoint. They measure variation across
continuations, not across independent initializations. \RepLimitations{}
The early anatomy-shaped result has incomplete continuation documentation,
and \textsc{anatomy-v2} was stopped after inspecting its epoch-75 deficit.
Its later trajectory is unknown. Probe optimization and initialization
can also affect  linear probe comparisons; the reported fixed-budget
protocol is not a measurement of the best achievable classifier for each
encoder. External foundation models such as RETFound under the same  linear
probe, other self-supervised objectives such as masked image modelling, and
target-selection rules derived from a formal objective are left for future
work.

The study uses one dataset and one downstream task. Strategies, checkpoints
and analyses were chosen after repeated inspection of the same test split,
although probe heads were fitted on training data and selected on validation
data. Reported intervals are therefore descriptive rather than confirmatory.
The additional continuations use the same test split and therefore do not provide an independent external evaluation.
Validation-split differences are reported alongside every per-seed test
difference (Appendix~\ref{app:replication}).
FairVision has no layer annotations, so guide accuracy cannot be measured on
it. On $\CRGuideVolumes$ stratified training volumes, the \ArmBest{} band
disagreed grossly with the MIRAGE envelope in $\CRFOne\%$ of B-scans, more often
in glaucoma and low-signal scans, and $\CREnvAnyFlag\%$ of envelopes raised a
plausibility flag (Appendix~\ref{app:guides}). These are agreement statistics
between two automatic guides, not anatomical accuracy. Only the specified
coverage setting was pretrained, so we identify neither an optimal coverage
level nor a mechanism for its AUC decline.

\paragraph{Broader impact and ethics.}
This study uses a public, de-identified retinal dataset under its release terms
and involves no new recruitment or data collection, and no model reported here
is clinically validated or intended for deployment (Appendix~\ref{app:ethics}).

\subsection{Future work}
\label{sec:future}

Our model encodes B-scans separately and combines their features through mean pooling, without learning interactions between slices. Exploratory attentive probes also showed overfitting. Future work could retain individual slice features while learning relationships between neighboring B-scans before pooling, using V-JEPA-style prediction or 3D OCT models. Segmentation of complete retinal layers across a volume could provide anatomical guidance.

\section{Conclusion}

We investigated where I-JEPA should predict in retinal OCT.
 Anatomy-guided rectangle placement improved  glaucoma AUC under linear probing in
the original runs, including a simple intensity-centroid strategy.
More anatomically specific targeting did not consistently improve these
results, and the implementations differ in more than target location,
including the context they leave visible.
Comparisons that match guidance, target sizes and visible context across
independent pretraining runs remain the next step.

\bibliographystyle{plainnat}
\bibliography{references}

\appendix

\section{Additional experimental details}
\label{app:repro}

\paragraph{Pretraining and linear evaluation.}
The input crop is $256\times256$, with patch size 16 and a ViT-Base encoder.
The predictor has depth 6 and hidden dimension 384.
Pretraining uses AdamW with peak learning rate $0.00025$, a five-epoch
warmup, final learning rate $0.000001$, weight decay scheduled from $0.04$
to $0.4$, and EMA coefficient from $0.996$ to $1.0$.
The effective batch size is 512, formed from batches of 64 with gradient
accumulation. Rectangle target scales are $0.15$--$0.20$ of image area,
aspect ratios $0.75$--$1.5$, and context scales $0.85$--$1.0$.
Random resized crops use area scale $0.3$--$1.0$.
The guidance probability increases over the epoch-25 to epoch-30
continuation window; fixed-length anatomy targets use $K=16$ throughout.
The frozen linear head uses AdamW, learning rate $0.0004$, weight decay
$0.05$, batch size 256, seed 42, and a 50-epoch budget with validation-AUC
selection and patience 15. Paired analyses rely on the documented shared split
and deterministic loader ordering; historical prediction files lack
independent case-ID manifests.

\paragraph{B-scan selection and pooling.}
Each FairVision volume stores 200 B-scans of $200\times200$ pixels. Pretraining
and evaluation use the same 100 B-scans per volume, with indices
$\lfloor 199k/99\rfloor$ for $k=0,\ldots,99$ (0, 2, \ldots, 196, 199), each
resized bilinearly to $256\times256$ pixels. Pretraining samples every B-scan
independently as a random resized crop. Evaluation encodes the full B-scans,
averages the 256 patch features of each B-scan and then the 100 B-scan vectors
(dimension 768) before the LayerNorm--linear head. The secondary pooling
variants use the patch maximum within B-scans, the maximum across B-scans, or
the concatenated mean and maximum across B-scans.

\paragraph{Guide sources.}
\textsc{envelope} uses a GOALS-tuned MIRAGE-Large model with a ConvNeXt segmentation decoder and four classes: Elsewhere, RNFL, GCIPL, and choroid. Elsewhere includes other retinal tissue and image background. The union of the three foreground classes is repaired across the unlabelled gap between the inner-retinal layers and choroid to form the hard retinal envelope.

The anatomy-shaped and coverage strategies use precomputed soft guides from MIRAGE-Base with a ConvNeXt decoder and harmonized GOALS, Duke DME, and AROI annotations. GOALS and Duke DME RNFL/GCIPL labels are combined as InnerRetina, and AROI's pre-combined inner-retinal band maps to the same class; choroid remains separate. The segmentation head emits four logits: Elsewhere, InnerRetina, choroid, and an ignored void channel. Softmax is applied over all four logits before retaining only the InnerRetina and choroid score maps for guidance. The \textsc{anatomy-v1} guide uses a decoder-feature adapter; \textsc{anatomy-v2} and \textsc{cover} use an encoder-feature adapter. The models and adapters are frozen when these guides are generated and cached. These guide products are not independently controlled in the cross-strategy comparison. \ArmBest{} uses image intensity rather than a
segmentation model: in the trained configuration its band
(Equation~\ref{eq:centroid}) is $\CRBandRows$ rows high and spans the central
$\CRBandCols$ columns of the patch grid (region fraction $\CRBandRegion$,
lateral fraction $\CRBandLateral$).
The adapters optimize feature-similarity objectives on FairVision-derived
image caches, not a glaucoma classification loss. The guide models and adapters remain frozen during I-JEPA training. The retained provenance does not independently establish
subject-level non-overlap for all data used to train the guidance models.

\paragraph{Included and excluded continuations.}
\label{app:excluded}
The early anatomy-v1 configuration resumes an \textsc{envelope} epoch-27
checkpoint rather than a documented epoch-25 fork, and its producing launch record
is incomplete. Its only result is at epoch 30, where it lies
$\DAnatomyOneEnvelopeEpThirty$ pp above \textsc{envelope} (CI
$\DAnatomyOneEnvelopeEpThirtyCI$). This comparison inherits the lineage
uncertainty.
Anatomy-v2 results use the continuation with full-precision target-encoder
outputs. Probes from its earlier mixed-target-precision continuation are
excluded, as are earlier uniform controls that used different target
precision and context selection. The main table retains the comparable
strategy endpoints, including the adverse \textsc{cover} trajectory.

\paragraph{Mask measurement.}
\label{app:geomprov}
Table~\ref{tab:geometry} replays the sampler code in each strategy's trained
configuration (commit {\footnotesize\texttt{\CRSamplerCommit}}), including
the \ArmBest{} band above and \textsc{envelope} placement on the hard envelope
with the fallback rule used in training. The scope is $\CRAuditVolumes$
training volumes with $\CRAuditSlices$ views each; every strategy sees the same
crops, block sizes and placement seeds. Tissue is guide occupancy of at least
$0.25$ on the soft MIRAGE envelope, on the $16\times16$ patch grid. Table~\ref{tab:delivered} summarizes per-image and batched geometry. In Table~\ref{tab:geometry},
``tissue hidden'' is the share of tissue cells in the target union, ``purity''
the share of target cells on tissue, ``unique targets'' the number of unique
target cells, ``loss slots'' the supervised target tokens including
duplicates, and ``context'' the share of the grid supplied to the context
encoder; in Table~\ref{tab:delivered}, ``mask ratio'' is the number of unique
target cells divided by $N$. Repeated target indices add loss slots, and the
anatomy-shaped and coverage strategies use different guide products from the
\textsc{envelope} run.

\begin{table}[ht]
\centering
\caption{Mask geometry under the trained configurations and the
protocol of Table~\ref{tab:geometry}. Per-image columns are measured one view
at a time ($\CRImgDraws$ single-view draws per seed), before batching;
the last column is the context delivered at batch size $\CRAuditBatch$.}
\label{tab:delivered}
\footnotesize
\setlength{\tabcolsep}{3pt}
\begin{tabular}{lcccccc}
\toprule
& \multicolumn{5}{c}{per image} & delivered \\
\cmidrule(lr){2-6}
strategy & tissue hidden & purity & mask ratio & context & loss slots & context \\
\midrule
\textsc{random}     & \CRImgHidRandom         & \CRImgPurRandom         & \CRImgMaskRandom         & \CRImgCtxRandom         & \CRImgSlotsRandom         & \CRCtxRandom \\
\ArmBest{}          & \CRImgHidCentroid     & \CRImgPurCentroid     & \CRImgMaskCentroid     & \CRImgCtxCentroid     & \CRImgSlotsCentroid     & \CRCtxCentroid \\
\textsc{envelope}   & \CRImgHidEnvelope     & \CRImgPurEnvelope     & \CRImgMaskEnvelope     & \CRImgCtxEnvelope     & \CRImgSlotsEnvelope     & \CRCtxEnvelope \\
\textsc{cover}      & \CRImgHidCover           & \CRImgPurCover           & \CRImgMaskCover           & \CRImgCtxCover           & \CRImgSlotsCover           & \CRCtxCover \\
\textsc{anatomy-v2} & \CRImgHidAnatomyTwo & \CRImgPurAnatomyTwo & \CRImgMaskAnatomyTwo & \CRImgCtxAnatomyTwo & \CRImgSlotsAnatomyTwo & \CRCtxAnatomyTwo \\
\bottomrule
\end{tabular}
\end{table}

\paragraph{Data use and ethics.}
\label{app:ethics}
This is secondary research on the public, de-identified FairVision release
under its CC BY-NC-ND 4.0 data licence. No new participants were recruited
or identifying fields added. A separate institutional ethics or exemption
determination is not recorded for this project. The models are not clinically
validated or intended for patient care. The illustrative OCT B-scan in
Figures~\ref{fig:target_selection} and~\ref{fig:policies} comes from the
separately attributed CC BY 4.0 OCTDL publication,
not from the FairVision evaluation cohort.

\section{Additional continuation and evaluation details}
\label{app:replication}
Two additional completed continuations each of \textsc{random}, \ArmBest{} and \textsc{envelope} start from the shared epoch-25 checkpoint. Seeds set data order, crops, mask draws and data-loader streams; runs with the same seed share data order only. \textsc{envelope} retains the original fallback when no admissible window has low overlap: a uniform choice among admissible windows. A uniform-placement control receives \ArmBest{}'s per-image budgets for seed A. Optimizer, schedule horizon and effective batch size match the original continuations, with full-precision teacher targets; each run stops at epoch~\RepEndpoint{}. The original \textsc{random} and \ArmBest{} continuations used different GPU hardware from the original \textsc{envelope} continuation and all additional runs.

Linear probing follows the main protocol in fp32, with \RepNHeadSeedsWord{} independently seeded linear heads fitted to each feature cache. We report their mean AUC and standard deviation; the original encoders are re-probed in the same way. Checkpoint decisions use validation AUC. All additional comparisons are descriptive. Differences from \textsc{random} of the same seed use paired test-case bootstrap intervals, which describe test-case sampling rather than independent pretraining runs. The pooling comparisons use seed A. Visual-field mean-deviation regression in glaucoma cases and validation-split differences are secondary analyses.

\RepAppendixResults

\begin{table}[t]
\centering
\caption{Every evaluated encoder at epoch~\RepEndpoint{}. AUC is the mean over \RepNHeadSeedsWord{} probe-head seeds, with their s.d. $\Delta$ (pp) is the difference from \textsc{random} of the same seed with a paired, label-stratified test-case bootstrap interval (\RepNBoot{} resamples), and the validation difference is given for comparison. Intervals describe test-case sampling, not retraining. All comparisons are descriptive.}
\label{tab:replication_runs}
\scriptsize
\setlength{\tabcolsep}{3pt}
\begin{tabular}{llccclcl}
\toprule
policy & run & test AUC & head s.d. & val.\ AUC & $\Delta$ (pp) test (interval) & $\Delta$ (pp) val. & analysis \\
\midrule
\textsc{random} & original & \RepAUCRandomOrig & \RepHeadSDRandomOrig & \RepValAUCRandomOrig & --- & --- & original \\
\textsc{random} & seed A & \RepAUCRandomSeedA & \RepHeadSDRandomSeedA & \RepValAUCRandomSeedA & --- & --- & descriptive \\
\textsc{random} & seed B & \RepAUCRandomSeedB & \RepHeadSDRandomSeedB & \RepValAUCRandomSeedB & --- & --- & descriptive \\
\ArmBest{} & original & \RepAUCCentroidOrig & \RepHeadSDCentroidOrig & \RepValAUCCentroidOrig & $\RepDeltaOrigCentroid$ & $\RepValDeltaOrigCentroid$ & original \\
\ArmBest{} & seed A & \RepAUCCentroidSeedA & \RepHeadSDCentroidSeedA & \RepValAUCCentroidSeedA & $\RepDeltaCentroidSeedA$ $\RepDeltaCentroidSeedACI$ & $\RepValDeltaCentroidSeedA$ & descriptive \\
\ArmBest{} & seed B & \RepAUCCentroidSeedB & \RepHeadSDCentroidSeedB & \RepValAUCCentroidSeedB & $\RepDeltaCentroidSeedB$ $\RepDeltaCentroidSeedBCI$ & $\RepValDeltaCentroidSeedB$ & descriptive \\
\textsc{envelope} & original & \RepAUCEnvelopeOrig & \RepHeadSDEnvelopeOrig & \RepValAUCEnvelopeOrig & $\RepDeltaOrigEnvelope$ & $\RepValDeltaOrigEnvelope$ & original \\
\textsc{envelope} & seed A & \RepAUCEnvelopeSeedA & \RepHeadSDEnvelopeSeedA & \RepValAUCEnvelopeSeedA & $\RepDeltaEnvelopeSeedA$ $\RepDeltaEnvelopeSeedACI$ & $\RepValDeltaEnvelopeSeedA$ & descriptive \\
\textsc{envelope} & seed B & \RepAUCEnvelopeSeedB & \RepHeadSDEnvelopeSeedB & \RepValAUCEnvelopeSeedB & $\RepDeltaEnvelopeSeedB$ $\RepDeltaEnvelopeSeedBCI$ & $\RepValDeltaEnvelopeSeedB$ & descriptive \\
\textsc{random-cb} & seed A & \RepAUCRandomCBSeedA & \RepHeadSDRandomCBSeedA & \RepValAUCRandomCBSeedA & $\RepBudgetShift$ $\RepBudgetShiftCI$ & $\RepValBudgetShift$ & control \\
\bottomrule
\end{tabular}
\end{table}

\begin{table}[t]
\centering
\caption{Pooling variants for seed A at epoch~\RepEndpoint{}. Pooling is applied to patch tokens within a B-scan, then across B-scans. Test AUC is averaged over \RepNHeadSeedsWord{} probe-head seeds; $\Delta$ (pp) is the difference from \textsc{random}. Comparisons are exploratory.}
\label{tab:pooling}
\scriptsize
\setlength{\tabcolsep}{3pt}
\begin{tabular}{lccccc}
\toprule
pooling & \textsc{random} & \ArmBest{} & \textsc{envelope} & $\Delta$ \ArmBest{} & $\Delta$ \textsc{envelope} \\
\midrule
mean, mean (primary) & \RepMeanAUCRandom & \RepMeanAUCCentroid & \RepMeanAUCEnvelope & $\RepDeltaMeanCentroid$ & $\RepDeltaMeanEnvelope$ \\
max, mean & \RepPoolPatchMaxAUCRandom & \RepPoolPatchMaxAUCCentroid & \RepPoolPatchMaxAUCEnvelope & $\RepPoolPatchMaxDeltaCentroid$ & $\RepPoolPatchMaxDeltaEnvelope$ \\
mean, max & \RepPoolSliceMaxAUCRandom & \RepPoolSliceMaxAUCCentroid & \RepPoolSliceMaxAUCEnvelope & $\RepPoolSliceMaxDeltaCentroid$ & $\RepPoolSliceMaxDeltaEnvelope$ \\
mean, mean and max & \RepPoolMeanMaxAUCRandom & \RepPoolMeanMaxAUCCentroid & \RepPoolMeanMaxAUCEnvelope & $\RepPoolMeanMaxDeltaCentroid$ & $\RepPoolMeanMaxDeltaEnvelope$ \\
\bottomrule
\end{tabular}
\end{table}

\begin{table}[t]
\centering
\caption{Visual-field mean deviation (MD, dB) regressed from frozen epoch-\RepEndpoint{} features in glaucoma cases of the test split ($n=\RepMDN$), ridge regression with the penalty chosen on validation. Lower MAE and RMSE and higher Spearman correlation are better. $\Delta$ is the difference from \textsc{random} of the same seed with a paired case-bootstrap interval. MD partly defines the glaucoma label, so this is a severity endpoint on the same cohort, not an independent task. Encoders without a per-case feature-cache identity (\textsc{random} original) are not included.}
\label{tab:md}
\scriptsize
\setlength{\tabcolsep}{3pt}
\begin{tabular}{llcccll}
\toprule
policy & run & MAE & RMSE & Spearman & $\Delta$ MAE (interval) & $\Delta$ Spearman \\
\midrule
\textsc{random} & seed A & \RepMDMAERandomSeedA & \RepMDRMSERandomSeedA & \RepMDSpearmanRandomSeedA & --- & --- \\
\textsc{random} & seed B & \RepMDMAERandomSeedB & \RepMDRMSERandomSeedB & \RepMDSpearmanRandomSeedB & --- & --- \\
\ArmBest{} & original & \RepMDMAECentroidOrig & \RepMDRMSECentroidOrig & \RepMDSpearmanCentroidOrig & --- & --- \\
\ArmBest{} & seed A & \RepMDMAECentroidSeedA & \RepMDRMSECentroidSeedA & \RepMDSpearmanCentroidSeedA & $\RepMDDeltaMAECentroidSeedA$ $\RepMDDeltaMAECentroidSeedACI$ & $\RepMDDeltaSpearmanCentroidSeedA$ \\
\ArmBest{} & seed B & \RepMDMAECentroidSeedB & \RepMDRMSECentroidSeedB & \RepMDSpearmanCentroidSeedB & $\RepMDDeltaMAECentroidSeedB$ $\RepMDDeltaMAECentroidSeedBCI$ & $\RepMDDeltaSpearmanCentroidSeedB$ \\
\textsc{envelope} & original & \RepMDMAEEnvelopeOrig & \RepMDRMSEEnvelopeOrig & \RepMDSpearmanEnvelopeOrig & --- & --- \\
\textsc{envelope} & seed A & \RepMDMAEEnvelopeSeedA & \RepMDRMSEEnvelopeSeedA & \RepMDSpearmanEnvelopeSeedA & $\RepMDDeltaMAEEnvelopeSeedA$ $\RepMDDeltaMAEEnvelopeSeedACI$ & $\RepMDDeltaSpearmanEnvelopeSeedA$ \\
\textsc{envelope} & seed B & \RepMDMAEEnvelopeSeedB & \RepMDRMSEEnvelopeSeedB & \RepMDSpearmanEnvelopeSeedB & --- & --- \\
\textsc{random-cb} & seed A & \RepMDMAERandomCBSeedA & \RepMDRMSERandomCBSeedA & \RepMDSpearmanRandomCBSeedA & $\RepMDDeltaMAERandomCBSeedA$ $\RepMDDeltaMAERandomCBSeedACI$ & $\RepMDDeltaSpearmanRandomCBSeedA$ \\
\midrule
\multicolumn{2}{l}{training-mean baseline} & \RepMDBaseMAE & \RepMDBaseRMSE & --- &  &  \\
\bottomrule
\end{tabular}
\end{table}

\begin{figure}[t]
\centering
\includegraphics[width=\linewidth]{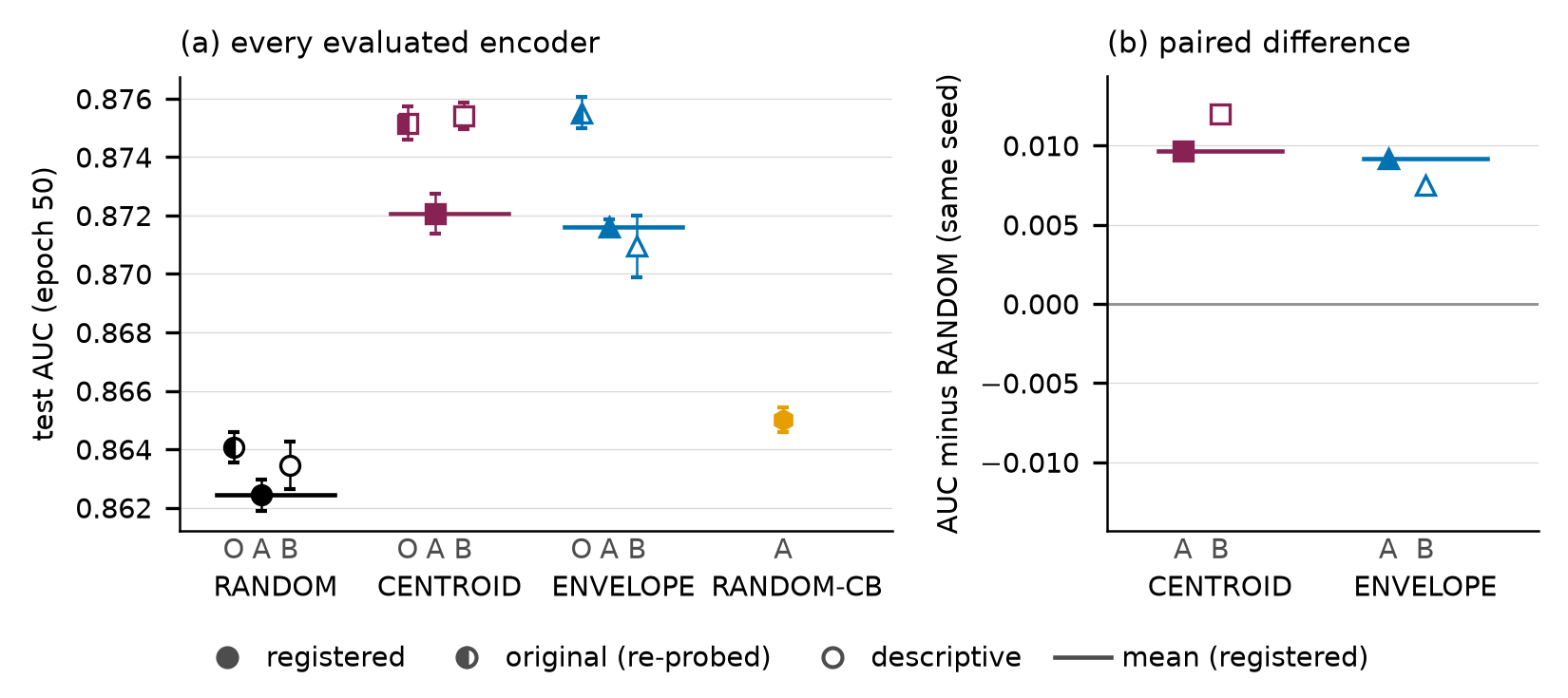}
\caption{ Variation across pretraining continuations at epoch~\RepEndpoint{}. (a) Test AUC averaged over \RepNHeadSeedsWord{} probe-head seeds; error bars show head-seed standard deviation. (b) Paired AUC differences (on the original AUC scale) from \textsc{random}. Bars summarize seed A. Filled and open markers distinguish additional continuations A and B; half-filled markers denote re-probed originals (O). Letters identify seeds. All continuations share an ancestor checkpoint.}
\label{fig:seed_spread}
\end{figure}

\section{Guide agreement and envelope plausibility}
\label{app:guides}
FairVision has no retinal layer annotations, so neither guide can be scored for
accuracy on it. We fixed failure rules before computing any rate and compared
the \ArmBest{} band with the hard MIRAGE envelope (patch occupancy at least
$0.25$) on full B-scans of $\CRGuideVolumes$ training volumes,
$\CRGuidePerStratum$ from each combination of glaucoma label and tertile of a
signal-to-noise proxy, with ten B-scans per volume ($\CRGuideBscans$ B-scans).
The computation was repeated on all $\CRCensusVolumes$ training volumes. A
B-scan counts as a gross disagreement when the band--envelope IoU is below
$0.2$ or more than half of the band cells contain no envelope. Overall rates
are population-weighted. Subgroup rates are unweighted proportions within the
stratified sample. Disagreement rates use B-scans with a usable envelope, and
intervals are 95\% volume-cluster bootstrap intervals.

The band disagreed grossly with the envelope in $\CRFOne\%$ of B-scans
(\CRFOneCI, all training volumes $\CRFOneCensus\%$) and was off-center in at
least three of ten columns in $\CRFTwo\%$. Disagreement was more frequent in
glaucoma than in normal eyes ($\CRFOneGlaucoma\%$ versus $\CRFOneNormal\%$) and
in the lowest than in the highest signal tertile ($\CRFOneLowSNR\%$ versus
$\CRFOneHighSNR\%$). The median IoU was $\CRIoUMedian$. Because the envelope is
typically about half as thick as the $\CRBandRows$-row band, an IoU of this
size is expected even for a centered band. Of the envelopes, $\CREnvAnyFlag\%$
(\CREnvAnyFlagCI) raised at least one plausibility flag: an empty or invalid
envelope, three or more components ($\CREnvFragmented\%$), irregular thickness
($\CREnvThickness\%$) or implausible position ($\CREnvPosition\%$). These are agreement and plausibility
statistics between two automatic guides, not anatomical accuracy. The envelope
model's only quantitative validation is in-domain on GOALS (foreground Dice
$\CRMirageDice$).

\section{Target sizing and visible context}
\label{app:covbug}

\paragraph{Batch equalization.}
Predictor targets are shortened to the minimum target length in the
microbatch, and the loss applies a single mean reduction to the stacked
prediction and target tensors, including averaging over images.
\textsc{cover} chooses placements using the anatomical mass
inside full candidate rectangles before this equalization.
Because target indices are ordered row-major, shortening removes the
lower portion of each rectangle rather than a spatially random subset.
Coverage at placement time can consequently differ from coverage at the loss.

\paragraph{Measured geometry.}
Across $24{,}000$ emitted targets, $73.4\%$ remain perfect rectangles.
In the batch-level audit of the trained configurations, \textsc{cover} hides
$\CRHidCover\%$ of tissue cells against $\CRHidEnvelope\%$ for \textsc{envelope}
while supplying less context (Table~\ref{tab:geometry}), and the guide mass
hidden at the loss is $\DTLegacyDeliveredMass\%$ against
$\DTLegacyScoredMass\%$ at placement (Table~\ref{tab:delivered_repair}).
Thus the trained \textsc{cover} policy is not a controlled higher-coverage
counterpart to \textsc{envelope}.

\textsc{envelope} shares the same shortening rule but does not optimize an
exact coverage level. The anatomy-shaped strategies instead resample
targets to a fixed $K{=}16$ indices. Equalization therefore has different
consequences across the strategies, even when nominal target counts agree.

\section{Final target and context masks}
\label{app:delivered_audit}

\paragraph{Scope.}
We replayed the actual guided dataset, paired transforms, DataLoader worker
collator and returned mask tensors on the existing 600-view training scope
(24 volumes, 25 views each), using fixed new crop draws shared across policies.
These are not reconstructions of the original training crops. Drawn block sizes were
injected independently of placement; equal seeds alone do not pair
policy-dependent rejection sampling. The production-size subset contains
\DTFullImages{} views in complete batches of 64, with \DTValidGuides{} valid
post-crop guides. The shorter final batch is excluded from
Table~\ref{tab:delivered_repair}. Views within volumes are correlated, so these
are descriptive mask statistics, not independent training replications.

\paragraph{What is measured.}
Target coverage is measured both at scoring and after target collation.
Encoder-visible tissue is counted only in the final encoder indices, not
in the full-grid target complement. Tissue is a guide occupancy thresholded
at $0.25$, not clinical ground truth. The context criterion is at least
$\max(\lceil 0.21N_{\mathrm{tissue}}\rceil,\min(4,N_{\mathrm{tissue}}))$
non-target tissue cells. This occupancy-cell diagnostic is distinct from
the policy's soft-mass constraint on the target complement.

\begin{table}[h]
\centering
\caption{Final masks from the trained \textsc{cover} strategy on
\DTFullImages{} fixed training views. Context-criterion misses count only
the \DTValidGuides{} valid guides. Scored and final coverage refer to
supported soft guide mass; context counts include all views.}
\label{tab:delivered_repair}
\small
\begin{tabular}{lr}
\toprule
Measurement & COVER \\
\midrule
Views losing tissue targets during equalization & \DTLegacyTruncation \\
Valid-guide context-criterion misses & \DTLegacyFloorMiss \\
Views with zero tissue context & \DTLegacyZeroContext \\
Scored hidden guide mass (\%) & \DTLegacyScoredMass \\
Final hidden guide mass (\%) & \DTLegacyDeliveredMass \\
Mean encoder-visible tissue cells & \DTLegacyContextMean \\
\bottomrule
\end{tabular}
\end{table}

\paragraph{Random fill is not a background quota.}
At full ramp all usable targets take the guided branch, while the
\texttt{random\_legal} fill mechanism is conditional on unused target blocks.
Some views have no random-fill blocks. Such blocks can include tissue, and
guided blocks can include background, so branch labels cannot stand in for
background supervision. Target composition is therefore measured from
the returned token indices rather than inferred from the branch name.

\section{Motivating attribution analysis and evidence limits}
\label{app:interp}

\paragraph{Origin of the target-selection question.}
Before studying anatomy-guided masking, we inspected feature-occlusion maps
from glaucoma classifiers initialized with the uniformly masked I-JEPA
baseline and then fine-tuned. The analysis compared mean pooling,
cross-attention pooling and an attentive probe.
At the B-scan level, a B-scan representation was zeroed before re-evaluating
the classifier. At the patch level, an encoded patch was omitted from its
B-scan's patch mean, the mean was recomputed from the remaining patches,
and the altered B-scan representation was passed to the pooling module
and classification head. The change in logit measures sensitivity to this
feature intervention; it is not pixel occlusion or an attention map.

Let $F$ denote the sequence of B-scan representations supplied to the
classifier and $\ell(F)$ its glaucoma logit. If $F^{(-s,p)}$ replaces
the representation of B-scan $s$ with the mean of its remaining patch
features after omitting patch $p$, the attribution is
\begin{equation}
 \Delta_{s,p}=\ell(F)-\ell\!\left(F^{(-s,p)}\right).
 \label{eq:occlusion_delta}
\end{equation}
Positive values mean that removing the feature lowers the glaucoma logit;
negative values mean that removal raises it. Both signs indicate
prediction sensitivity. Selected archived displays are not reproduced because
of the dataset licence.

The historical report and selected overlays prompted us to investigate retinal
regions as prediction targets. Those examples were selected from correctly
classified cases and informative B-scans, so they are not a population-level
estimate of anatomical localization. Nor does a response to feature removal
establish a clinically causal region or show which pretraining mask is best.
Fine-tuning used diagnosis labels, so the maps cannot separate the contribution
of self-supervised pretraining from subsequent supervised adaptation.

\paragraph{Available evidence.}
Archived occlusion curves and heat maps remain available, but the underlying
per-volume numerical arrays and a complete normalization record are not
retained in the evidence used for this paper. The historical record also
documents sensitivity of patch-level displays to numerical precision and
colour scaling. The selected image panels are retained as archived
visualizations, without restoring their numerical summaries as validated
results. We make no spatial-consistency, laterality or clinical
landmark claim from those displays. Recovering the producing arrays,
checkpoint identities and plotting transforms would be necessary for a
reproducible quantitative analysis.

\section{Full paired-contrast table}
\label{app:contrasts}

\begin{table}[t]
\centering
\caption{Original-run contrasts: matched epoch \emph{and} matched precision. $\Delta$
is a paired difference on identical test cases; the interval is a
\Nboot-resample paired bootstrap and $p$ is a DeLong test for correlated ROC
curves. $q$ is Benjamini--Hochberg over these nine contrasts. All $p$ and $q$
values here are descriptive (Section~\ref{sec:limits}).
Every comparison against the null excludes zero.}
\label{tab:contrasts}
\small
\begin{tabular}{llcccc}
\toprule
contrast & epoch & $\Delta$ (pp) & 95\% CI (pp) & $p$ & $q$ \\
\midrule
\textsc{envelope} $-$ \textsc{random} & 50  & \DEnvelopeRandomEpFifty  & \DEnvelopeRandomEpFiftyCI  & \DEnvelopeRandomEpFiftyP  & \DEnvelopeRandomEpFiftyQ \\
\textsc{envelope} $-$ \textsc{random} & 75  & \DEnvelopeRandomEpSeventyFive  & \DEnvelopeRandomEpSeventyFiveCI  & \DEnvelopeRandomEpSeventyFiveP  & \DEnvelopeRandomEpSeventyFiveQ \\
\textsc{envelope} $-$ \textsc{random} & 100 & \DEnvelopeRandomEpHundred & \DEnvelopeRandomEpHundredCI & \DEnvelopeRandomEpHundredP & \DEnvelopeRandomEpHundredQ \\
\addlinespace
\ArmBest{} $-$ \textsc{random}   & 50  & \DOracleRandomEpFifty    & \DOracleRandomEpFiftyCI    & \DOracleRandomEpFiftyP    & \DOracleRandomEpFiftyQ \\
\ArmBest{} $-$ \textsc{random}   & 75  & \DOracleRandomEpSeventyFive    & \DOracleRandomEpSeventyFiveCI    & \DOracleRandomEpSeventyFiveP    & \DOracleRandomEpSeventyFiveQ \\
\ArmBest{} $-$ \textsc{random}   & 100 & \DOracleRandomEpHundred   & \DOracleRandomEpHundredCI   & \DOracleRandomEpHundredP   & \DOracleRandomEpHundredQ \\
\addlinespace
\ArmBest{} $-$ \textsc{envelope} & 50  & \DOracleEnvelopeEpFifty  & \DOracleEnvelopeEpFiftyCI  & \DOracleEnvelopeEpFiftyP  & \DOracleEnvelopeEpFiftyQ \\
\ArmBest{} $-$ \textsc{envelope} & 75  & \DOracleEnvelopeEpSeventyFive  & \DOracleEnvelopeEpSeventyFiveCI  & \DOracleEnvelopeEpSeventyFiveP  & \DOracleEnvelopeEpSeventyFiveQ \\
\ArmBest{} $-$ \textsc{envelope} & 100 & \DOracleEnvelopeEpHundred & \DOracleEnvelopeEpHundredCI & \DOracleEnvelopeEpHundredP & \DOracleEnvelopeEpHundredQ \\
\bottomrule
\end{tabular}
\end{table}

Benjamini--Hochberg is applied within a declared family consisting
of exactly these nine contrasts (the fp16 comparisons among \textsc{random},
\ArmBest{} and \textsc{envelope} at epochs 50, 75 and 100) and the remaining
25 exploratory contrasts in the stored statistics are corrected as a separate
family. The two families are never pooled, and the $q$ values printed above are
those of the nine-contrast family only. Declaring the family fixes what is
corrected together. It does not make these contrasts confirmatory, because the
test split was inspected repeatedly before the family was fixed, and the paper
reports them as descriptive throughout.

\section{Probe precision control}
\label{app:fp32}

The original rectangle-strategy probes use mixed precision, while the
anatomy-shaped and coverage-strategy probes use fp32. All $\Delta$ values in
Table~\ref{tab:main} use a baseline matched in both epoch and probe precision.
Probe precision is distinct from pretraining target-encoder precision.
We repeated the recorded mixed-precision probes with autocast disabled in
feature extraction, head fitting and evaluation. Each comparison uses the
same frozen encoder. The recorded changes are smaller than the principal
strategy differences.

\begin{table}[ht]
\centering
\caption{Full fp32 re-probe against the original fp16 result, same frozen
encoder in each row. The eight DeLong $p$-values are unadjusted diagnostics, not
a family-wise test. Differences in this table use the original AUC scale. The largest difference is $2\times10^{-4}$.}
\label{tab:fp32}
\small
\begin{tabular}{lccccc}
\toprule
policy & epoch & fp16 AUC & fp32 AUC & $\Delta$ & $p$ \\
\midrule
\textsc{envelope} & 50 & 0.876064 & 0.876063 & -0.000001 & 0.964 \\
\textsc{envelope} & 75 & 0.880307 & 0.880305 & -0.000002 & 0.806 \\
\textsc{envelope} & 100 & 0.880743 & 0.880761 & +0.000018 & 0.167 \\
\ArmBest{} & 50 & 0.874030 & 0.874015 & -0.000015 & 0.412 \\
\ArmBest{} & 100 & 0.885485 & 0.885293 & -0.000192 & 0.767 \\
\textsc{random} & 50 & 0.864097 & 0.864121 & +0.000024 & 0.128 \\
\textsc{random} & 75 & 0.872302 & 0.872302 & -0.000001 & 0.962 \\
\textsc{random} & 100 & 0.874581 & 0.874485 & -0.000096 & $<$0.001 \\
\bottomrule
\end{tabular}

\end{table}

\end{document}